\documentclass[preprint,10pt]{elsarticle}

\usepackage{amssymb}
\usepackage{amsmath}

\usepackage{lineno}

\usepackage[T1]{fontenc}

\DeclareGraphicsExtensions{.pdf,.jpeg,.jpg,.png}
\graphicspath{{./figs/}}
\usepackage{rotating}
\usepackage{algorithm}
\usepackage{algorithmicx}
\usepackage{algpseudocode}
\usepackage{color}

\usepackage{eqparbox}

\journal{Applied Soft Computing}

\begin{document}

\begin{frontmatter}



\title{Classification and Feature Transformation with Fuzzy Cognitive Maps}


\author{Piotr Szwed\corref{cor1}}
\ead{pszwed@agh.edu.pl}
\address{AGH Univeristy of Science and Technology, Krak\'ow, Poland}

\begin{abstract}
Fuzzy Cognitive Maps (FCMs) are considered a soft computing technique combining elements of fuzzy logic and recurrent neural networks. 
They found multiple application in such domains as modeling of system behavior, prediction of time series, decision making and process control. Less attention, however, has been turned towards using them in pattern classification. 
In this work we propose an FCM based classifier with  a fully connected map structure. In contrast to methods that expect reaching a steady system state during reasoning, we chose to execute a few FCM iterations (steps) before collecting output labels.  Weights were learned with a gradient  algorithm and logloss or cross-entropy were used as the cost function.
Our primary goal was to verify, whether such  design would result in a descent general purpose classifier, with performance comparable to off the shelf classical methods.   
As the preliminary results were promising, we investigated the hypothesis that the performance of $d$-step classifier can be attributed to a fact that in previous $d-1$ steps it transforms the feature space by grouping observations belonging to a given class, so that they became more compact and separable. To verify this hypothesis we calculated three clustering scores for the transformed feature space. We also evaluated performance of pipelines built from FCM-based data transformer followed by a classification algorithm. The standard statistical analyzes confirmed both the performance of FCM based classifier and its capability to improve data. The supporting prototype software was implemented in Python using TensorFlow library. 
\end{abstract}


\end{frontmatter}


\section{Introduction}
\label{sec:intro}

Fuzzy Cognitive Maps (FCMs) consist  of concepts  linked by causal relations. Concepts model features, events or conditions appearing in a analyzed domain. Causal relations between concepts are expressed by directed edges with assigned real valued weights, which describe strength of positive or negative influence between concepts. 
FCMs have their roots in cognitive maps that were first proposed by Axelrod \cite{Axelrod1976} as a tool for modeling political decisions; next, they were extended by Kosko \cite{Kosko1986,Kosko1992} by introducing fuzzy concept activation levels.

FCMs are considered a soft computing technique with appealing graphical form, which facilitates interpretability. Their models constitute a knowledge representation capturing domain information with selection of concepts and edge weights. FCMs can be either entered by domain experts or learned from data with a data-driven \cite{stach2008data} algorithm. 

FCMs are executable, after setting initial concept activation levels a trajectory of system states can be computed using a simple state equation. Due to generality of the model that allows backward connections and cycles they are also referred as a ``kind of recurrent neural networks'' \cite{napoles2018fuzzy}.   

During last 30 years numerous applications of FCMs have been proposed in various domains.
Their (not exhaustive) list includes:  
support of medical decisions \cite{papageorgiou2008brain}, identifying gene regulatory network from gene expression data \cite{chen2015inferring}, recognition of emotions  in speech \cite{zhang2017new},
process control \cite{papageorgiou2004active}, 
ecosystem modeling \cite{Ozesmi2004}, analysis of development of economic systems and  introduction of new technologies \cite{Jetter2011}, academic units development \cite{SzwedFCM2013}, 
prediction of time series \cite{stach2008data}, traffic prediction \cite{ChmielSzwed2015}
project risk modeling \cite{Lazzerini2011}, reliability engineering \cite{SALMERON20123818} and security risk assessment for IT systems 
\cite{SzwedSkrzynski14,SzwedSkrzynskiChmiel14}.  

Majority  of research related to FCMs was focusd on such problems as: modeling, decision making  and prediction of dynamic behavior, however, as stated in the recent review \cite{napoles2018fuzzy} ``less attention has been given to the development of FCM-based classifiers''.  

Actually, several classifiers using internally FCMs has been proposed, although their number is relatively small. Moreover, in many cases they were tested on propriety data or transformed datasets bounded to a particular domain, hence, comparison with other methods would be hard.  Only in a few cases they were validated using publicly available datasets from UCI repository. 

Basically, development of an FCM classifier involves four steps: (i) establish a map structure,
(ii) select a method of feeding the model with data, (iii) choose how to collect classification output form the map,
(iv) select a cost function and an optimization algorithm to learn model parameters from data
The above design decisions can be combined in various ways yielding a plethora of solutions. 

In this work we investigated an FCM based classifier with  a fully connected map structure allowing backward links between input and output concepts. In contrast to methods that expect reaching a steady system state, we choose to execute a few FCM iterations (steps) before collecting output labels. The number of iterations is a model parameter, referred further as \emph{depth}. Weights were learned with a gradient  algorithm and logloss or cross-entropy were used as the cost function. It should be mentioned, that in the context of FCMs gradient methods were used mainly for time series prediction and mean square error was a typical choice of cost function, even for classification.

Our primary goal was to verify, whether such  design would result in a descent general purpose classifier, with performance comparable to off the shelf classical methods offered by various machine learning libraries.   

As the preliminary results were promising, we investigated the hypothesis that the performance of $d$-step classifier can be attributed to a fact that in $d-1$ steps it transforms the feature space in such a way, that the final step, which is equivalent to the Logistic Regression, is easier, i.e. it groups observations belonging to a given class so that they became more compact and separable. 
We expected that, if such behavior was confirmed, trained FCM classifier could be used as a preprocessing step improving data before delivering them to the actual classifier and in a lucky case the whole pipeline might outperform both: FCM and final classifier. 

To verify this hypothesis we calculated three clustering scores for the transformed feature space, namely: Davies-Bouldin, Silhouette Coefficient and Calinski-Harabasz. We also evaluated performance of pipelines built from FCM-based data transformer followed by selected classification algorithms from scikit-learn library. The standard statistical analyzes confirmed both the performance of FCM based classifier and its capability to improve data.

A prototype software was implemented in Python language using TensorFlow library. All computations were performed in a prototyping mode called \emph{eager execution} using intensively the offered option of symbolic gradient calculation, which is apparently slower than a hard-coded backpropagation algorithm. Nevertheless, we provide a suitable algorithm dedicated to FCMs that, although not integrated with the actual implementation, was tested to give identical results as symbolic calculation.
  
%

The rest of the paper is organized as follows: the next Section~\ref{sec:related_works}  surveys research related to FCM learning and applications in classification; Section~\ref{sec:fcm-classifier} introduces the FCM based classifier: its architecture and learning algorithm. Section~\ref{sec:experiments} defines the experimental setup, it is followed by Section~\ref{sec:results} providing results of experiments and their statistical analyzes. Finally, Section~\ref{sec:conclusion} gives concluding remarks.

\section{Related works}
\label{sec:related_works}

In this section we introduce mathematical formulation of Fuzzy Cognitive Maps model and discuss related research with a particular focus on application of FCMs in prediction and classification, as well as learning algorithms.
For a comprehensive view on FCMs and their applications we refer the reader to the surveys that were co conducted in 2005: \cite{Aguilar2005}, 2011: 2013: \cite{papageorgiou2013review} and 2017: \cite{felix2017review}. We would also like to mention a review on FCM learning algorithms  \cite{Papageorgiou2011} and applications of FCMs in pattern classification \cite{napoles2018fuzzy}.


\subsection{FCM model}
\label{subsec:fcm}

Fuzzy Cognitive Maps are graphs of concepts $C=\{c_1,\dots,c_n\}$ linked by causal relations expressed as directed edges with assigned weights. In the simplest FCM form weights indicating negative or positive influence have values from the set $\{-1, 1\}$, however, more often they take discrete or continuous values from the interval $[-1,1]$. FCM models can be built automatically or developed by experts.

Being a weighted directed graph, FCM can be represented by $n\times n$ influence matrix $W$, whose 
elements $w_{ij}$ comprise weights assigned to edges linking concepts $c_j$ and $c_i$. 
Weights equal to 0 indicate that there is no causal relation between related concepts.	
%
%
%
%
%
%

System state is a vector of concept activation levels. Usually they are bounded to lie within a selected interval, e.g. $[0,1]$ or $[-1,1]$. FCM execution consists in building a sequence of states: $\alpha =  A^{(0)}, A^{(1)},\dots, A^{(k)},\dots$ 
starting from an initial state $A^{(0)}$. Consecutive elements are calculated according to a recurrent state equation. In this work we refer to a general state equation given by  formula (\ref{eq:state-equation}). During $k$-th iteration the vector $A^{(k)}$ is multiplied by the weights matrix $W$ and its elements are brought to  the assumed range with \emph{activation} (or \emph{squashing}) function $f$.  

\begin{equation}
\label{eq:state-equation}
A_i^{(k+1)} = f\left(\sum\limits_{j=1}^n w_{ij}\,A_j^{(k)}\right)
\end{equation}

Depending on the intended range of activation levels ($[0, 1]$ or $[-1, 1]$) various activation functions can be used \cite{Bueno2009}, however, a very popular choice for the range $[0,1]$ is the sigmoid function given by formula (\ref{eq:sigmoid-fun}). 

\begin{equation}
\label{eq:sigmoid-fun}
f(x)=\frac{1}{1+\exp(-\lambda x)}
\end{equation}

In FCM literature several variants of state equations were formulated. The most salient difference is related to the presence of diagonal elements in the weights matrix $W$. They correspond to self-loops, i.e. influence relation linking past and current concept activation levels. Another variant of state equation uses the transposition of $W$.

\begin{figure}[ht!]
	{\scriptsize
		\begin{tabular}{ccc}
			\includegraphics[width=0.3\linewidth]{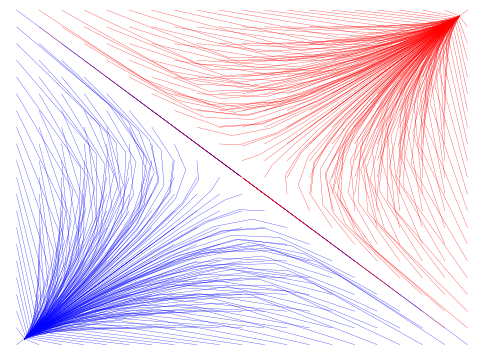}&
			\includegraphics[width=0.3\linewidth]{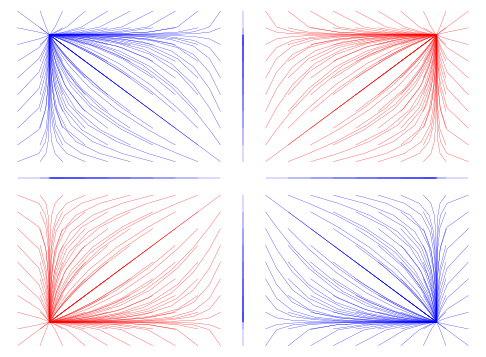}&
			\includegraphics[width=0.3\linewidth]{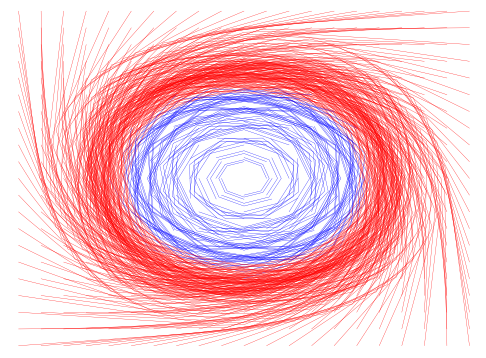}\\
			(a)&(b)&(c)
		\end{tabular}
	}
	\caption{ Examples of FCM attractors:  (a) two fixed point (b) four fixed point (c) cyclic. Trajectories of 2D FCMs starting at evenly spaced initial points tend to concentrate approaching attractors. Separate attraction basins are marked with different colors.}
	\label{fig:ex_fcm_attractors}
\end{figure}

The weights matrix and a type of activation function determine the FCM behavior: after a finite number of iterations the execution trajectory  reaches a steady state, which is either a fixed point, a cyclic or a chaotic attractor.  Fig.~\ref{fig:ex_fcm_attractors} shows three examples of FCM attractors in 2D space (a) two fixed points, (b) four fixed points and (c) a cyclic one.

FCM trajectory can be interpreted in two ways:
\begin{itemize}
	\item As a representation of a dynamic behavior of the modeled system. Consecutive states are linked with an implicit temporal relation and the whole sequence describes the system evolution over time.  
	Hence, developed FCM model can be used to predict future system states and for example be applied for analysis of evolution scenarios or time series prediction.   
	\item The sequence of states may be treated as a fuzzy inference process, in which selected elements of a steady state comprise reasoning results. This approach gives an opportunity to use FCMs for such tasks as decision making or classification.  In particular fixed point attractors can serve this purpose. They may be attributed with a resulting decision, whereas their attraction basins can be considered decision regions. 
	
\end{itemize}

\subsection{Learning}
\label{subsec:learning}

The goal of FCM learning is to find a set of weights assuring desired map behavior during prediction or inference. In    
\cite{felix2017review} two main groups of learning algorithms were distinguished: Hebbian and and error driven. 

Methods based on Hebb's rule  iteratively update weights based on observed simultaneous changes of concept activation levels.  
Typically, Hebbian methods take as input the map structure, an initial set of weights established by experts and ranges of desired values of output concepts. 
Hebbian learning was applied in construction of one of the first FCM-based classifiers  \cite{papakostas2012towards}.

Error-driven methods use a cost function expressing differences (errors) between predicted and expected FCM outputs. They apply  optimization techniques  to find  the model parameters, which minimize the cost.  The survey \cite{felix2017review} gives a good summary of various cost functions used by particular learning methods. 
They are built upon terms expressing absolute or squared errors computed for final map output or for sequences of states generated during a few FCM iterations. Sequences of activation levels are taken into account for time series prediction tasks \cite{stach2007parallel} but also in pattern classification, where they are referred by \emph{convergence error} \cite{napoles2018fuzzy}.  

In some cases cost functions included penalties for concept values beyond the expected ranges, as well as Lasso L1 regularization factors \cite{chen2015inferring,WuLiuLiuShen2020}.  It can be observed, that even for classification tasks, the selection of cost functions was typical for prediction of continuous outputs (regression).


Over years several optimization methods were applied to learn FCMs. Among them the most popular were Genetic Algorithm (GA), Real Coded Genetic Algorithm (RCGA), Particle Swarm Optimization (PSO), Differential Evolution (DE), Ant Colony Optimization (ACO) and Simulated Annealing (SA).
An exhaustive list of references can be found in two surveys \cite{Papageorgiou2011,felix2017review}.

Population algorithms seem to prevail among error-driven learning methods. In the majority of cases they were used to learn small FCMs comprising about a dozen concepts. Unless special restrictions are introduced to enforce sparsity, as is the case of certain FCM architectures used for classification \cite{papakostas2010classifying}, the weights matrices are dense and the number of their elements grows quadratically with the number of concepts. This constitute a serious performance challenge for population optimization methods, in which each solution within a population represents a full weights matrix.  

In the last few years a number of applications of gradient-based methods to learn FCM  have been  reported. As computation of gradients requires the cost function to be differentiable, usually mean squared error was selected. In \cite{Madeiro2012} gradient based local search was used to improve RCGA and DE. Pure gradient method relaying on backpropagation trough time (BPTT) was described in \cite{gregor2013training}. 
Similar approach was taken in \cite{gregor2013tuning} to tune single FCM attractor position. Multistep gradient based algorithms turned out to be successful in time series prediction \cite{jastriebow2014analysis,poczeta2014analysis,papageorgiou2017fuzzy}. The algorithm implemented a gradient based rule for weights updating, which takes into account multiple observations in the past. 
Also a gradient algorithm was applied in large scale online FCM learning for time series prediction \cite{WuLiuLiuShen2020}.

\subsection{Convergence of sigmoid function}
\label{subsec:conv_sig_fun}
Depending on the application of FCM, it may be desirable that a map has multiple stable fixed point attractors.
This in particular pertains to such tasks as classification, for which predicted class labels can be assigned to attractors and decision regions to attraction basins (c.f. Fig.~\ref{fig:ex_fcm_attractors} ). Construction of classifiers based on such approach was presented in \cite{papakostas2008fuzzy,papakostas2010classifying,napoles2014improve,NAPOLES2016154,froelich2017towards}.

In \cite{knight2014linear} authors derived a condition of existence of a single stable fixed point for sigmoid activation function given by equation (\ref{eq:sigmoid-fun}), which is one of the most often used for FCMs. For a given map size $n$ (i.e. the number of concepts), the number of fixed points depends on value of $\lambda$, the slope parameter. Small values of $\lambda$ guarantee uniqueness of a fixed point; on the contrary, large values, above a certain boundary value $\lambda(n)$ result in multiple fixed points. For $n=2$ the boundary value $\lambda(2)=2$ and  $\lambda(n)$ quickly decays to 0 as $n$ grows. Hence, application of FCMs in classification tasks requires setting larger $\lambda$ values. 

\subsection{Classification}
\label{subsec:classification}

In survey \cite{napoles2018fuzzy}  two groups of  FCM based classifiers were distinguished: \emph{low-level}, in which concepts represent input features   directly and \emph{high-level} based on abstract concepts and information granules. 

As the classifier proposed in this work can be considered  low-level, we will discuss solutions pertaining to the first group in more detail.   
Low-level FCM classifiers are composed from input and output concepts. Hidden concepts (neurons) are usually not present, as it would contradict the postulate  of interpretability \cite{napoles2018fuzzy}. Moreover, an architecture with hidden concepts was investigated in \cite{papakostas2012towards}, but it turned out to be less efficient than the others.  

During classification of a data instance $x$, its attributes are used to initialize the input concepts and then, after a number of FCM iterations,  a predicted class label can be established by analyzing activation level of one or more output concepts. 

The number of output concepts may vary. 
\begin{itemize}
\item In the FCM architecture referred in \cite{papakostas2012towards} and as \emph{class-per-output} each class is represented by a single concept. Such form corresponds to \emph{one-hot} encoding. 
\item In the case of \emph{single-output} architecture only one output concept is used regardless of the number of classes. In consequence, the range of activation levels of this single concept must be partitioned to regions corresponding to class labels. 
This can be done either with \emph{By Threshold}  (ByT) or with \emph{Class of Minimum Distance} – CoMD mappings. ByT method consist in selecting boundary values separating classes, whereas CoMD determines centers of activation level for a given class during the training phase and assigns class labels based on the nearest center.
\end{itemize}

FCMs were used as standalone classifiers or in combination with other methods. 
Configurations, in which object features and/or labels originating from another classifier were delivered to an FCM were analyzed in  \cite{papakostas2008fuzzy} and \cite{papakostas2010classifying}.
They were tested on a few datasets including synthetic ones, dataset from UCI repository and  image data.
Ensamble of FCM classifiers (following bagging and boosting approaches) were proposed in \cite{papageorgiou2012fuzzy} and applied to autism detection. 

In \cite{papakostas2012towards}  three variants of FCM architecture, referred jointly as \emph{FCMper},  were compared by performing tests on 8 datasets from UCI repository. A common feature of all investigated map structures were feed-forward connections responsible for mapping activation levels from input  to output concepts.  Main differences between FCMper variants were related to presence of connections within input and output layers. Surprisingly, the simplest feed forward architecture corresponding to two-layers perceptron turned out to be the most efficient. The same architecture combined with single-output and CoMD mapping was investigated in \cite{froelich2017towards}. The resulting classifier was tested on 15 datasets from UCI repository and compared with other machine learning methods.

As discussed earlier, classification methods learn map weights keeping the activation function fixed.  The method called Stability based on Sigmoid Function (SSF) \cite{napoles2014improve,NAPOLES2016154} is based on a different approach. It assumes that activation functions may vary, whereas FCM weights proposed by experts remain constant. Such solution is viable: if the employed sigmoid activation functions have slope parameters $\lambda$ large enough, the map can converge to multiple fixed point attractors \cite{knight2014linear} corresponding to class labels. 
An advantage of this method is that it preserves interpretability of the initial map. Moreover, the weights matrix developed by experts is usually sparse,  whereas the matrix resulting from the learning process tends to be dense.
The method using single output concept and PSO as the learning algorithm was tested on six real datasets characterizing resistance mechanism of HIV-1 proteins to existing drugs.


Rough Cognitive Networks (RCN) \cite{NAPOLES2016_46} are an interesting recent development of \emph{high-level} classifiers combining Rough Sets Theory and FCMs . 
Given a relation linking similar (indiscernable) data objects, a rough set partitions the feature space into three disjont regions: comprising objects that surely, possibly and certainly not belong to the set. During RCN learning  a rough set is created for each label  and its partition regions are established based on similarity relation. Finally, these regions are represented by three FCM  concepts.
A clear advantage of RCN over low level FCM classifiers  is the limited number of concepts: considering additional outputs, their total number is equal $4 \times$ number of class labels. 
In \cite{NAPOLES2017_79} authors decided to expand the network architecture introducing ensembles of RCN at different levels of granularity. A certain drawback of RCN was the need to tune parameters governing the similarity relation and recompute partitions. The problem was resolved in further extension named Fuzzy-Rough Cognitive Networks \cite{NAPOLES2018_19}.  

It should be mentioned that all RCN classifiers were extensively tested on a large pool of 140 datasets from UCI repository and showed performance comparable to best classifiers from Weka library.

\section{FCM Classifier}
\label{sec:fcm-classifier}

In this section we present a setup of the proposed classifier. We discuss the links between the last classification step and Logistic Regression and analyze the classifier properties on a small 2D  classification problem. Finally, we present the learning algorithm.

\subsection{Design}

During design of FCM based classifier several decision have been made. They are discussed in the subsequent paragraphs.

\paragraph{Depth} 

A characteristic feature of the proposed method is that it does not exploit trajectory convergence. 
Instead, a parameter \emph{depth}  corresponding to the number of FCM iterations is introduced.
Hence, for depth  $d$ trajectories have $d+1$ elements: $A^{(0)}, A^{(1)},\dots, A^{(d)}$. The best classification results are usually obtained for small values of $d$ (like 2,3,4). The depth is a  hyperparameter that can be established during a typical model selection process.

\paragraph{Feature range} As a feature range the interval $[0,1]$ was selected , i.e. all features are initially scaled to fit within this interval. 
As an activation function a slightly modified sigmoid given by formula (\ref{eq:activation-fun}) is used. 

\begin{equation}
\label{eq:activation-fun}
f(x)=\frac{1}{1+\exp(-\lambda(x-0.5))}
\end{equation}

The function maps $\mathbb{R} \to (0,1)$ and  is shifted along $X$ axis, so that its fixed point lays at 0.5, i.e. exactly at the middle of its range.
For an output variable the constant 0.5 can be interpreted as \emph{undecided} value, in between 0 and 1.  

The definition of activation function is extended to matrices and vectors, i.e. for $z \in \mathbb{R}^r\times \mathbb{R}^m$ ($z \in \mathbb{R}^r$), 
we will denote by $f(z)$ a matrix  $[f(z_{ij})]_{\substack{i=1,r\\j=1,m}}$ (or a vector $[f(z_i)]_{i=1,r}$).

\paragraph{FCM architecture}
Classifier can be considered a function $c:X \to C$ that maps feature space $X$ into a set of labels $C=\{c_1,\dots,c_k\}$. Assuming that $X = \mathbb{R}^n$, all $n$ input features are transformed into concepts.  FCM comprises also 1 to $k$ \emph{output concepts}, whose activation levels reached in final states are used for prediction. The number of output concepts depends on the problem type. For \emph{binary} classification, where $|C|=2$ one output concept can be used to determine two predicted labels.  For \emph{multiclass} problem consisting of $k$ labels, \emph{one hot} encoding is applied, i.e. 
each class is represented by a single output concept. 

As it is discussed further, the assumed architecture determines the type of loss function used during classifier fitting, therefore we treat them as separate configurations referenced as FCMBinaryClassifier (FCMB) and FCMMulticlassClassifier (FCMMC). It should be noted, that FCMMC can be also used for binary classification, moreover, it often outperforms FCMB in binary classification tasks.

Output concepts are not only collectors of class labels. In opposition to feed forward FCM architectures proposed in \cite{papakostas2008fuzzy,papakostas2010classifying,papakostas2012towards} or neural networks,  their values influence states of other concepts. In fact, they are nonlinear features computed from subsequently transformed input. Adding extra features can be  beneficial in some classification problems, as observations that are hard to discriminate in original feature space are more likely to be separated in the space of higher dimension.

\begin{figure}[ht]
{\centering
\scriptsize
\begin{tabular}{cc}
\includegraphics[scale=.5]{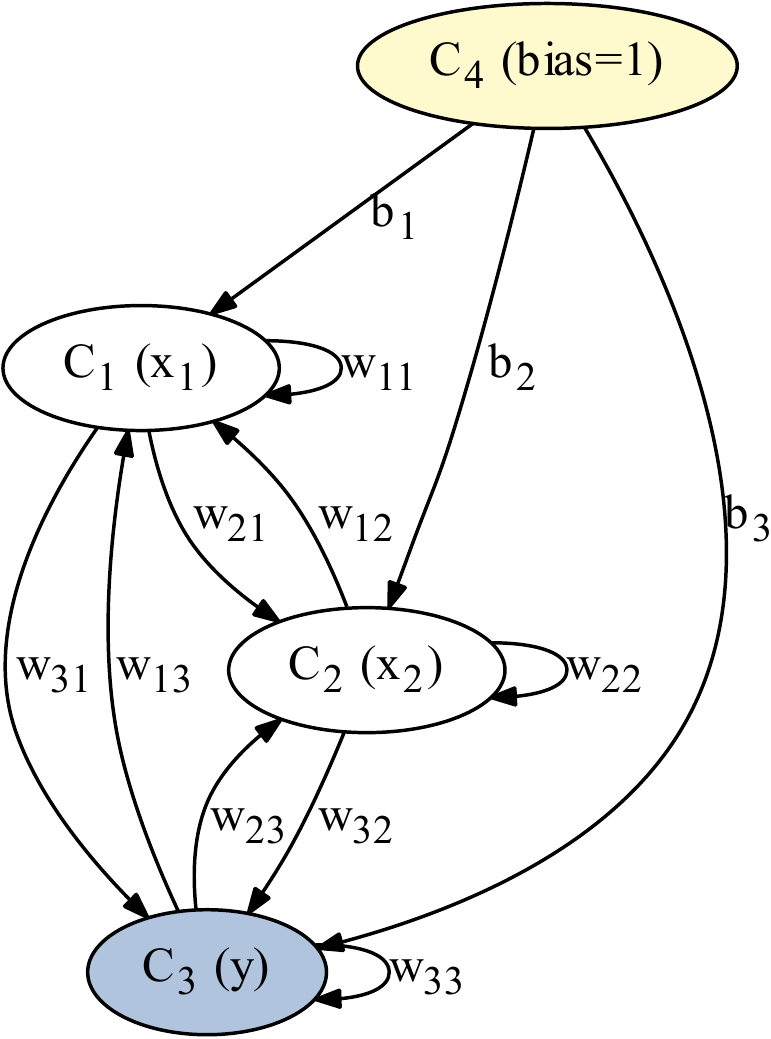}&
\includegraphics[scale=.5]{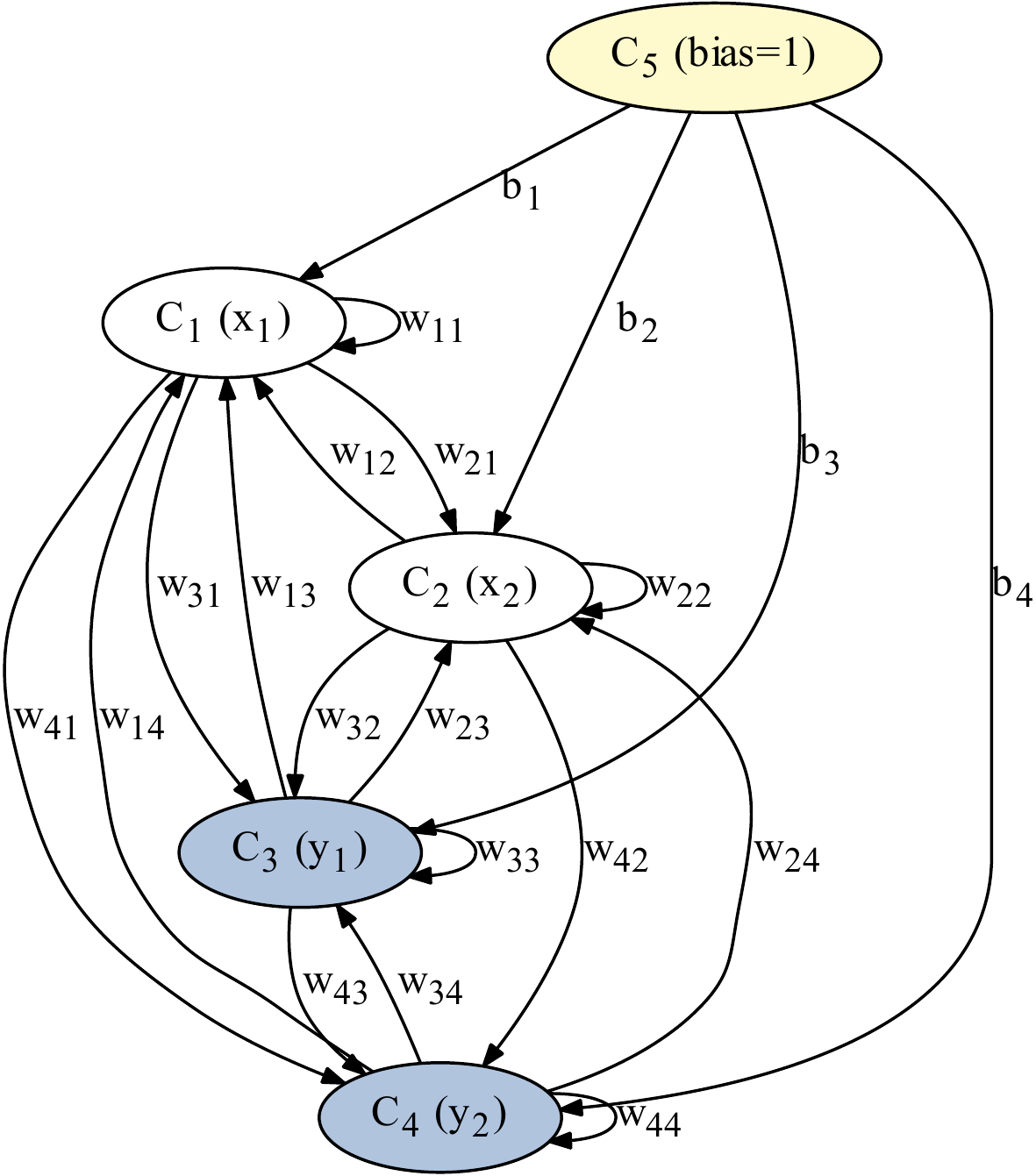}\\
(a)&(b)
\end{tabular}
}
\caption{Two FCM architectures for 2D binary classification problems  (a) FCMB -- single output concept (b) FCMMC -- two output concepts, each for class labels}
\label{fig:ex_fcm_architectures}
\end{figure}

Fig.~\ref{fig:ex_fcm_architectures} presents two alternative FCM configurations for 2-dimensional binary classification task. For FCMB one output concept ($C_3$) is added, whereas FCMMC the configuration comprises two output concepts: $C_3$ and $C_4$.   

The state vector $\mathrm{A}^{(t)}=\left[\begin{smallmatrix}\mathrm{x}^{(t)}\\\mathrm{y}^{(t)}\end{smallmatrix}\right]$ is a concatenation of vectors $\mathrm{x}^{(t)}\in \mathbb(R)^n$ for input concepts and $\mathrm{y}^{(t)}$ for output ones. Let us denote by $r$ the size of state vector $\mathrm{A}^{(t)}$.
Size of  $\mathrm{y}^{(t)}$ depends on the configuration: for FCMB $y^{(t)}\in \mathbb{R}$ and $r=n+1$, whereas for FCMMC $y^{(t)}\in \mathbb{R}^k$,  where $k$ is the number of class labels; in this case $r=n+k$. 

Activation levels of output concepts must be initialized before starting FCM execution. All elements of $\mathrm{y}^{(0)}$ are set to 0.5, what corresponds to undecided value.

\paragraph{State equation and the bias term}
We use an extended variant of FCM state equation (\ref{eq:state-eq-classifier}) comprising a bias term $\mathrm{b}\in \mathbb{R}^r$.  

\begin{equation}
\mathrm{A}^{(t+1)}=f(\mathrm{W}\cdot \mathrm{A}^{(t)} + \mathrm{b})
\label{eq:state-eq-classifier}
\end{equation}

The bias corresponds to a concept, whose value is permanently set to 1 during execution (c.f. Fig.~\ref{fig:ex_fcm_architectures}). Its role can be seen in two ways. First, even simple 2D experiments show that bias strongly influences routes of FCM trajectories and in consequence the shapes of decision regions.  Second, presence of bias facilitates distinction of observations belonging to different classes, in particular in neural networks and methods based on a hyperplane separating decision regions.

\paragraph{Prediction}

Let $x \in \mathbf{R}^n$ be an input observation for the classifier. 
Classification with FCM consists of three steps:
\begin{enumerate}
\item Encode $x$ into initial state vector $\mathrm{A}^{(0)}=\left[\begin{smallmatrix}\mathrm{x}\\\mathrm{y}^{(0)}\end{smallmatrix}\right]$. All elements of $\mathrm{y}^{(0)}$ are set to 0.5.     
\item Generate a finite sequence of states $\mathrm{A}^{(0)},\mathrm{A}^{(1)},\dots, \mathrm{A}^{(d)}$ using equation (\ref{eq:state-eq-classifier})
\item Extract predicted label $\hat{y}$ from $\mathrm{y}^{(d)}$, the state of output concepts from the last state vector in the sequence. 
\end{enumerate}

For FCMB  $\mathrm{y}^{(d)}$ is a scalar, hence, the predicted label is selected using equation (\ref{eq:y_pred_binary})
\begin{equation}
 \hat{y} =
  \begin{cases}
    0&\mbox{if }  \mathrm{y}^{(d)}<0.5 \\
    1& \mbox{if }  \mathrm{y}^{(d)}\ge 0.5
  \end{cases}
\label{eq:y_pred_binary}
\end{equation} 

In the case of FCMMC the predicted label corresponds to the index of maximal value in the vector $\mathrm{y^{(d)}}$. However, its elements are earlier normalized with \emph{softmax} function $\sigma(\mathrm{y})\colon \mathbb{R}^k \to \mathbb{R}^k$ given by (\ref{eq:sotmax})

\begin{equation}
\sigma(\mathrm{y})_j=\frac{\exp(\mathrm{y}_j)}{\sum_{i=1}^k \exp(\mathrm{y}_i)}, \text{for } j=1,\dots,k 
\label{eq:sotmax}
\end{equation} 

Finally, the predicted label is extracted as:

\begin{equation}
 \hat{y} = \mathrm{arg} \max \sigma(\mathrm{y}^{(d)}) 
\label{eq:y_pred_multiclass}
\end{equation} 

Actually, normalization with softmax is not required to correctly classify instances. The label can be predicted based on maximal element value.   We used softmax, which is differentiable, in order to apply gradient based learning method.

\subsection{Simple case where d=1}
Let us discuss properties of FCMB (the binary classifier) for a simple case, where the $d$ (depth) parameter is set to 1. 
It can be expected that even after one iteration the classifier should be capable of predicting class labels.
If $d=1$, the FCM trajectories  contain just two states: $\mathrm{A}^{(0)}$ and $\mathrm{A}^{(1)}$. State vector $\mathrm{A}^{(0)}$ has two components: $\mathrm{x}^{(0)}$ -- data to be classified and $\mathrm{y}^{(0)}$ (actually a scalar) set to 0.5.  The component  $\mathrm{y}^{(1)}$ within $\mathrm{A}^{(1)}$ represents an output, which can be further used to predict class label according to (\ref{eq:y_pred_binary}). 

As it can be noticed, values of $\mathrm{x}^{(1)}$ are not used.  Value of $\mathrm{y}^{(1)}$ depends only on the weights comprised in the last $r$-th row of $W$ and $b$, namely  $W_{r1},\dots,W_{rr}$ and $b_r$.

\begin{equation}
\mathrm{y}^{(1)} = f(W_{r1}\mathrm{x}^{(0)}_{1}+W_{r2}\mathrm{x}^{(0)}_{2}+\dots+W_{rn}\mathrm{x}^{(0)}_{n}+W_{rr}\mathrm{y}^{(0)} + \mathrm{b}_r) 
\label{eq:d_1_binary_y}
\end{equation}

Considering the definition of activation function given by (\ref{eq:activation-fun}), equation (\ref{eq:d_1_binary_y}) can be transformed to: 

\begin{equation}
\mathrm{y}^{(1)} = \frac{1}{1+\exp(-(\mathrm{w}'\mathrm{x}^{(0)}+\mathrm{b}'))}\\
\label{eq:d_1_binary_y_logreg}
\end{equation}
where: 
\begin{equation*}
\begin{array}{l}
\mathrm{w}'=\lambda[W_{r1},W_{r2},\dots,W_{rn}]\\
\mathrm{b}'=\lambda(0.5 W_{rr}+\mathrm{b}_r -0.5)
\end{array}
\end{equation*}

Formula (\ref{eq:d_1_binary_y_logreg}) is a well known definition of label probability $p(\hat{y}=1|\mathrm{x})$ in Logistic Regression \cite{larose2006data}. The expression $z(\mathrm{x}^{(0)})=\mathrm{w}'\mathrm{x}^{(0)}+\mathrm{b}'$ defines the class boundary as a hyperplane separating two regions. Value of $\mathrm{y}^{(1)}$ at the boundary is equal to 0.5, and tends to 0 or 1 for points  $\mathrm{x}^{(0)}$ placed far from the hyperplane. 
Analysis of this specific case explains, why the bias term was introduced. It allows to shift the hyperplane position, so that the observations are separated correctly.  

Analogous mechanism is expected for FCM based classifier with a depth parameter $d>1$. In this case the system trajectory 
\begin{equation*}
\alpha = (\mathrm{A}^{(0)},\dots, \mathrm{A}^{(d-1)},\mathrm{A}^{(d)})  
\end{equation*}
comprises $d+1$ elements  and the final computation of $\mathrm{y}^{(d)}$ occurs in the last step using 
transformed features $\mathrm{A}^{(d-1)}=[\mathrm{x}^{(d-1)},\mathrm{y}^{(d-1)}]^T$ according to the following equation:
\begin{equation*}
\mathrm{y}^{(d)} = f(W_{r1}\mathrm{x}^{(d-1)}_{1}+W_{r2}\mathrm{x}^{(d-1)}_{2}+\dots+W_{rn}\mathrm{x}^{(d-1)}_{n}+W_{rr}\mathrm{y}^{(d-1)} + \mathrm{b}_r) 
\label{eq:d_1plus_binary_y}
\end{equation*}

Possible advantages of classification with FCM over Logistic Regression can be attributed to two phenomena:
\begin{itemize}
\item A new feature $\mathrm{y}^{(d-1)}$  is introduced, it is a nonlinear function of original features $\mathrm{x}^{(0)}$. The additional feature facilitates finding a hyperplane separating observations
\item Original observations $\mathrm{x}^{(0)}$ are transformed into $\mathrm{x}^{(d-1)}$. It can be expected that this transformation of the feature space makes observations more condensed, arranges them into groups or places on a manifold, where they are easier to separate.    
\end{itemize}

\subsection{Example}
\label{sec:fcm-classifier:example}

In this section we present an example of binary classification performed  with FCMB and FCMMC classifiers. As the input a synthetic 2D dataset  generated with {\tt make\_moons} function of scikit-learn library was used. 

Parameters of both models are given in  Table~\ref{tab:ex_parameters}; for a moment, we do not discuss, how they were learned. 

\begin{table}[ht!]
\centering
{\scriptsize
\begin{tabular}{|c|c|}
\hline\centering
FCMB&FCMMC\\
\hline
\begin{minipage}{0.43\textwidth}
\begin{equation*}
W=
\begin{bmatrix}
    0.28 & -0.31 & -0.09 \\
    1.17 &  0.45  & -0.66 \\
    \mathbf{-2.43} &  \mathbf{3.65} & \mathbf{-1.92}
\end{bmatrix}
\end{equation*}
\end{minipage}
&
\begin{minipage}{0.5\textwidth}
\begin{equation*}
W=\begin{bmatrix}
2.89 & -1.50 & -0.29 & -1.01\\
5.77 & -1.43 & 5.61 & -4.42\\
\mathbf{3.31} & \mathbf{-6.80} & \mathbf{0.96} & \mathbf{0.75}\\
\mathbf{5.03} & \mathbf{6.75} & \mathbf{-1.02} & \mathbf{-0.46}
\end{bmatrix}
\end{equation*}
\end{minipage}\\
\vspace{3pt}
\begin{minipage}{0.43\textwidth}
\begin{equation*}
b=
\begin{bmatrix}
0.28\\
0.57\\
\mathbf{-1.62}
\end{bmatrix}
\end{equation*}
\end{minipage}
&
\begin{minipage}{0.5\textwidth}
\begin{equation*}
b=
\begin{bmatrix}
-3.14\\
-1.38\\
\mathbf{3.01}\\
\mathbf{-2.18}
\end{bmatrix}
\end{equation*}
\end{minipage}
\\
$\lambda=5$, $d=3$ &
$\lambda=2$, $d=3$
\\
\hline
\end{tabular}
}
\caption{Parameters of FCMB and FCMMC classifiers}
\label{tab:ex_parameters}
\end{table}

\begin{figure}[ht!]
{\scriptsize
\begin{tabular}{cc}
\includegraphics[width=0.5\linewidth]{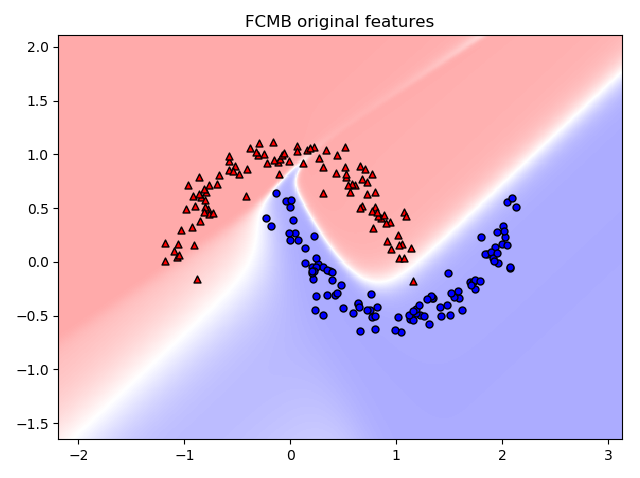}&
\includegraphics[width=0.5\linewidth]{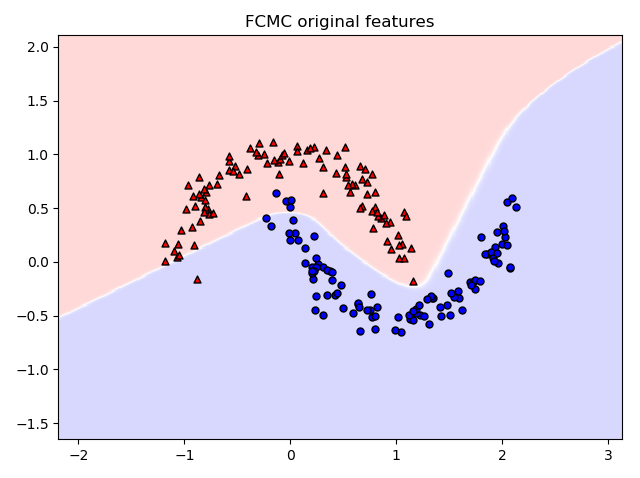}\\
(a)&(b)
\end{tabular}
}
\caption{Example data set and class boundaries (a) FCMB classifier (b) FCMMC classifier}
\label{fig:ex_original_feature_space}
\end{figure}

Fig.~\ref{fig:ex_original_feature_space} shows the data points as well as decision regions for both classes. Color intensity indicates 
the value returned by the test expressions appearing in equations (\ref{eq:y_pred_binary}) and (\ref{eq:y_pred_multiclass}), 
i.e. $|\mathrm{y}^{(1)}-0.5|$ for FCMB and $\max\{|\mathrm{y}^{(1)}_1-0.5|, |\mathrm{y}^{(1)}_2-0.5|\}$ for FCMMC. Both expressions are often interpreted as conditional probability $P(y=c_j|x)$ of returning class label $c_j$ for an observation $x$ laying in a given point. 
Although for FCMB all observations fit within correct decision regions, the shape of class boundary is more complex, when compared with FCMMC. This may suggest that FCMMC has better generalization properties (what was confirmed experimentally in the majority of cases).

\begin{figure}[ht!]
{\scriptsize
\begin{tabular}{cc}
\includegraphics[width=0.5\linewidth]{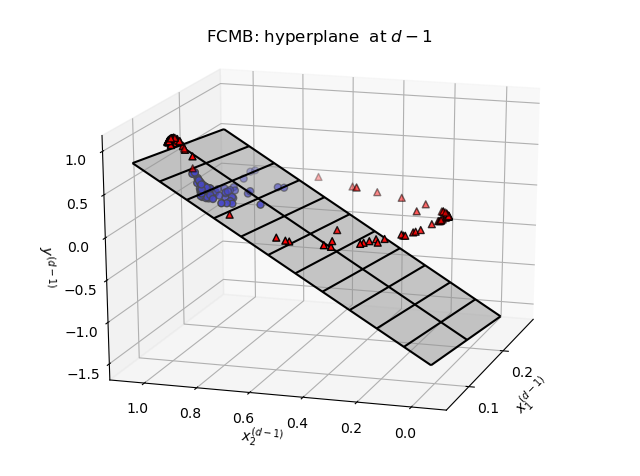}&
\includegraphics[width=0.5\linewidth]{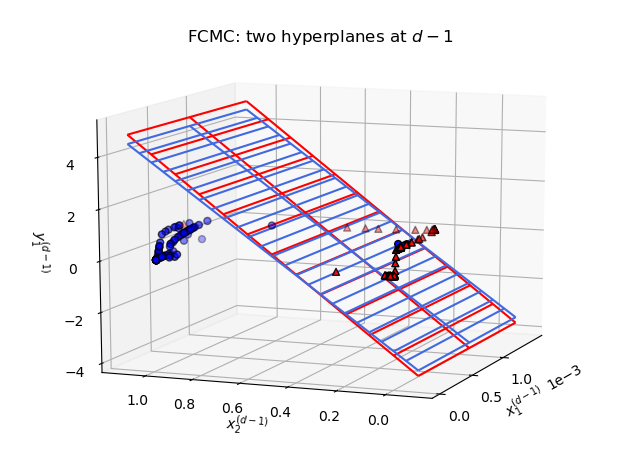}\\
(a)&(b)
\end{tabular}
}
\caption{3D projection of transformed feature space at $d-1$  (a) FCMB classifier - one hyperplane separating data; blue points are placed below the hyperplane and red points above (b) FCMC classifier - two hyperplanes, each for a class label. Coordinates of hyperplanes were calculated for $y_*^{(d-1)}=0.5$ }
\label{fig:ex_hyperplane_transformed}
\end{figure}

Fig.~\ref{fig:ex_hyperplane_transformed} presents locations of state vectors $A^{(d-1)}$ after $d-1$ iterations for FCMB and FCMMC classifier configurations.
It shows also hyperplanes separating transformed observations. Parametres of hyperplanes are marked in Table~\ref{tab:ex_parameters} with boldface. As it can be observed, FCMB employs only one hyperplane, whereas for FCMMC two close hyperplanes are defined. (In general, for FCMMC the number of hyperplanes correspond to the number of classes.)

\begin{figure}[ht!]
{\scriptsize
\begin{tabular}{cc}
\includegraphics[width=0.5\linewidth]{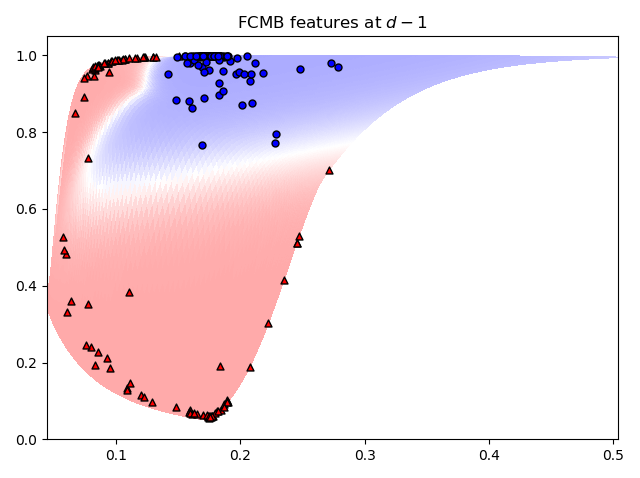}&
\includegraphics[width=0.5\linewidth]{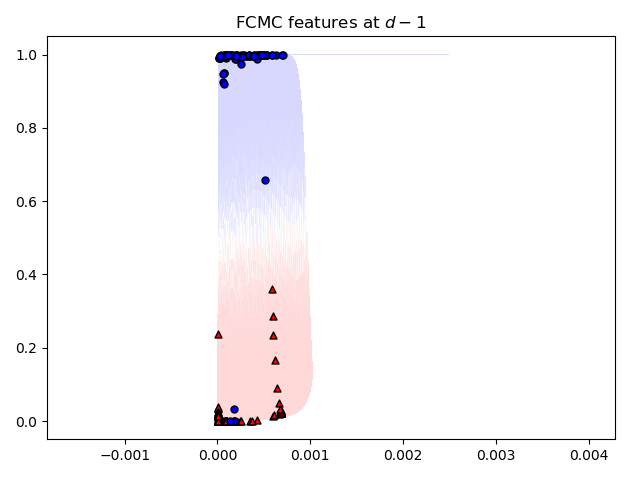}\\
(a)&(b)
\end{tabular}
}
\caption{Transformed feature space at $d-1$ and class boundaries (a) FCMB classifier (b) FCMC classifier}
\label{fig:ex_transformed_feature_space}
\end{figure}

Fig.~\ref{fig:ex_transformed_feature_space} shows 2D views of transformed feature space at $d-1$. It can be noticed that observations belonging to the same class are rearranged, so as they form groups or are placed on lower dimensional manifold. This in particular concerns the situation depicted in Fig.~\ref{fig:ex_transformed_feature_space}b, where  $0 \leq x_1^{(d-1)}\leq 0.001$ and $0 \leq x_2^{(d-1)}\leq 1$.


Based on the above examples we may formulate a hypothesis that FCM  classifier firstly acts as supervised feature transformer. It shifts the observations and places them in regions, where they would be easier to separate with a hyperplane (or multiple hyperplanes in the case of FCMMC) in the last step.

\subsection{Learning}

A typical approach for learning model parameters consist in defining an objective function $J$ dependent both on observations in the dataset and model parameters and then finding the model parameters, for which $J$ attains its optimal value. Close relations between the last step of FCM based classifier and   Logistic Regression suggest using Maximum Likelihood estimate, in particular log loss for FCMB or cross-entropy for FCMMC.

Let $G\colon \mathbb{R}^n \times \Theta \to \mathbb{R}^{r-n}$ be a function that maps input observations $x \in \mathbb{R}^n$ and model parameters $\Theta$ into set of activation levels of output concepts after  $d$ iterations. In this case model parameters correspond to weights in FCM state equation (\ref{eq:state-eq-classifier}), i.e. $\Theta = (\mathrm{W},\mathrm{b})$. The $\lambda$ parameter of the activation function in (\ref{eq:activation-fun}) and depth $d$ are considered hyperparameters to be established during model selection process.

Computation of  $\tilde{y} = G(x,\mathrm{W},\mathrm{b})$ is carried out in a number of steps given by formula (\ref{eq:classifier_function}). 

\begin{equation}
\begin{array}{ll}
\mathrm{A}^{(0)}&=encode(x)\\
\mathrm{A}^{(1)}&=f(\mathrm{W} \cdot {A}^{(0)} + \mathrm{b})\\
\dots\\
\mathrm{A}^{(d-1)}&=f(\mathrm{W} \cdot {A}^{(d-2)} + \mathrm{b})\\
\mathrm{A}^{(d)}&=f(\mathrm{W} \cdot {A}^{(d-1)} + \mathrm{b})\\
\tilde{y} &=extract(\mathrm{A}^{(d)})\\
\end{array}
\label{eq:classifier_function}
\end{equation}

\noindent First, an input observation $x$ is encoded into the initial state vector with function $encode\colon \mathbb{R}^n\to\mathbb{R}^r$ defined as:

\begin{equation*}
encode(x)= \left[\begin{smallmatrix}x\\\mathrm{y}\end{smallmatrix}\right], \text{where}\, \mathrm{y}=[0.5]_{i=1,r-n}
\end{equation*}

\noindent Next, the state equation (\ref{eq:state-eq-classifier}) is applied $d$ times and finally, activation levels of output concepts are selected with function $extract\colon \mathbb{R}^r\to\mathbb{R}^{r-n}$ function: 

\begin{equation*}
extract(A)= [A_i]_{i=n+1,r}
\end{equation*}

\noindent Function $G(x,\mathrm{W},\mathrm{b})$ is clearly a composed function that can be rewritten as:

\begin{equation}
G(x,\mathrm{W},\mathrm{b}) = extract(\underbrace{f(\dots f}_{d \, \text{times}}(encode(x)\cdot\mathrm{W}+\mathrm{b})\dots)\cdot\mathrm{W}+\mathrm{b}))
\end{equation}

\subsubsection{Cost function}
Let $D$ be a training dataset comprising $m$ pairs of observations $x_i$ and their labels $y_i$: $D=\{(x_i,y_i)\colon x_i\in \mathbb{R}^n,y_i\in C\}_{i=1,m}$.

We denote by $\tilde{y_i}$ the classifier output for $x_i$, hence $\tilde{y_i}=G(x_i,\mathrm{W},\mathrm{b})$. The symbol $\tilde{y_i}$ is used to distinguish continuous FCM output being a scalar for FCMB or a vector for FCMMC from a predicted discrete label $\hat{y}$ assigned with formulas  (\ref{eq:y_pred_binary}) or (\ref{eq:y_pred_multiclass}).  Further, let $\tilde{y}=(\tilde{y}_i)_{i=1,m}$ be a sequence of classifier outputs and $y=(y_i)_{i=1,m}$ a corresponding sequence of ground truth labels.

For FCMB, which intended exclusively for binary classification, we use log loss is as a cost function. Assuming that the set of class labels $C=\{0,1\}$ the cost function for $\tilde{y}$ and $y$ is defined according to (\ref{eq:logloss}) as:

\begin{equation}
\mathcal{L}_{FCMB}(\tilde{y},y) = 
-\frac{1}{m}\sum_{i=1}^m \left[ y_i \log(\tilde{y}_i) + (1-y_i)\log(1-\tilde{y_i})\right] 
\label{eq:logloss}
\end{equation}

In the case of FCMMC there are $k$ labels and the classifier function $G(x_i,\mathrm{W},\mathrm{b})$ returns vectors in $\tilde{y}_i \in\mathbb{R}^k$. As during prediction outputs are normalized with softmax function $\sigma$ defined by (\ref{eq:sotmax}), the 
same procedure should be applied before calculating the loss. Hence, the loss with respect to $\tilde{y}$ and $y$ is defined as cross-entropy preceded by softmax normalization:

\begin{equation}
\mathcal{L}_{FCMMC}(\tilde{y},y) = \\
-\frac{1}{m}\sum_{i=1}^m \sum_{j=1}^k  \mathbb{I}(y_i=c_j) \log(\sigma(\tilde{y}_i)_j) 
\label{eq:softmaxcrossentropy}
\end{equation}

By the term $\mathbb{I}(cond)$ in (\ref{eq:softmaxcrossentropy}) we denote indicator function, which returns 1 if condition $cond$ holds and 0 otherwise.

Finally, to calculate cost function for FCMB and FCMMC, the expression $G(x_i,\mathrm{W},\mathrm{b})$ should be plug into equations (\ref{eq:logloss}) and  (\ref{eq:softmaxcrossentropy}) in place of $\tilde{y}_i$, what gives:
 
\begin{multline}
J_{FCMB}(D,\mathrm{W},\mathrm{b}) =\\ 
-\frac{1}{m}\sum_{i=1}^m \left[ y_i \log(G(x_i,\mathrm{W},\mathrm{b})) + (1-y_i)\log(1-G(x_i,\mathrm{W},\mathrm{b}))\right]\\ 
\label{eq:logloss_cost}
\end{multline}

\begin{equation}
J_{FCMMC}(D,\mathrm{W},\mathrm{b}) = 
-\frac{1}{m}\sum_{i=1}^m \sum_{j=1}^k  \mathbb{I}(y_i=c_j) \log(\sigma(G(x_i,\mathrm{W},\mathrm{b}))_j) 
\label{eq:softmaxcrossentropy_cost}
\end{equation}

The goal of a learning algorithm is to find model parameters $(\mathrm{W}^*,\mathrm{b}^*)$ minimizing the cost function for a given a dataset $D$:
\begin{equation}
(\mathrm{W}^*,\mathrm{b}^*) = \mathrm{arg} \min_{\mathrm{W},\mathrm{b}} J(D,\mathrm{W},\mathrm{b})
\end{equation}

%

\subsubsection{Gradient based learning algorithm}

Learning FCM model is carried out with a standard gradient  algorithm, which in each step computes gradient of the cost function $J(D,\mathrm{W},\mathrm{b})$ with respect to $\mathrm{W}$ and $\mathrm{b}$ and then updates the weights. Depending on the algorithm, the modification applied to model parameters is a function of current and past gradient values scaled by a small constant (the learning rate). 

The algorithm pseudocode is given in Algorithm~\ref{alg:learning}. We will briefly discuss its parameters and the main steps. $D$ is an input dataset used for training. Parameter $d$ (depth) defines the number of FCM iterations. In each iteration (c.f. line 11) new FCM state is calculated with equation (\ref{eq:classifier_function}).

Following the terminology used for neural networks, an epoch is a single processing step over the whole dataset. The parameter $epochs$ defines their number.  The input dataset can be divided into disjoint subsets (batches) $B =\{ B_s\}_{s=1,t}$ that are processed separately. Random partitioning into batches is repeated in each epoch. The model parameters are updated after each batch, what converts the algorithm into a stochastic gradient descent type. The size of each batch is controlled by the parameters $bs$.  The total number of batches is equal $\lfloor \frac{m}{bs} \rfloor$.
Finally, $lr$ is the learning rate. 

\newdimen{\algindent}
\setlength\algindent{1.5em}          
\algnewcommand\LeftComment[2]{%
	\hspace{#1\algindent}$\triangleright$ \eqparbox{}{#2} \hfill %
}

\begin{algorithm}
\caption{Gradient based FCM learning algorithm} \label{alg:learning}
\begin{algorithmic}[1]
\Require
\Statex $D=\{(x_i,y_i)\}_{i=1,m}$ - input dataset
\Statex $d$ -- depth (number of FCM iterations)
\Statex $bs$ -- batch size
\Statex $epochs$ -- number of epochs
\Statex $lr$ -- learning rate
\State Initialize $W$ and $b$ with random values
\For{$t = 1$ \textbf{to} $epochs$ }
	\State Randomly divide $D$ into $t=\lfloor \frac{m}{bs} \rfloor$ batches: $B =\{ B_s\}_{s=1,t}$ 
	\ForAll{$B_s \in B$ }
		\Statex \LeftComment{2}{Start GradientTape}
		\State $\tilde{y} = ()$
		\State $y = ()$
		\ForAll{$(x_i,y_i)\in B_s$} 
			\State $A^{(0)} = encode(x_{i})$
			\State Perform $d$ FCM iterations for staring from $A^{(0)}$:
			\newline \hspace*{6em}$A^{(d)} \gets FCM(A^{(0)},W,b,d)$
			\State $\tilde{y}_i \gets extract(A^{(d)})$
			\State $\tilde{y} \gets \tilde{y} \circ (\tilde{y}_i)$ \Comment Add $\tilde{y}_i$ to the sequence $\tilde{y}$
			\State $y \gets y \circ (y_i)$ \Comment Add $y_i$ to the sequence $y$
		\EndFor	
		\State Compute current loss: $l=\mathcal{L}(\tilde{y},y)$
		\Statex \LeftComment{2}{Stop GradientTape}
		\State $(\nabla_W J,\nabla_b J) = compute\_gradient(\mathcal{L}(\tilde{y},y),\mathrm{W},\mathrm{b})$	
		\State Update weights: $(W,b) = update(W,b,\nabla_W J,\nabla_b J lr)$ 
	\EndFor	
\EndFor
\end{algorithmic}
\end{algorithm}

\subsubsection{Implementation}

The algorithm was implemented in Python language using TensorFlow library, which contains a number of  functions useful for learning neural networks, including partitioning into batches and several gradient based algorithms for updating model parameters (c.f. Algorithm~\ref{alg:learning}:17).
Tensorflow is integrated with Keras library, which in turn offers higher level abstractions, e.g. various types of neural network layers, which can be used as building blocks for construction of complex models. However, they do not comprise a component that would mimic FCM behavior.

Typically, TensorFlow (and Keras) models are compiled into computational graphs that can be efficiently executed on CPU or GPUs from NVidia. However, the framework offers a mode designed for prototyping and debugging called \emph{eager execution}. A unique feature of eager execution is \emph{GradientTape}. With GradientTape it is possible to record a sequence of  function calls, next perform symbolic differentiation with respect to selected set of variables, and finally compute gradients, by substituting variable symbols  in the symbolic derivative with actual values.

This mechanism was actually applied in the prototype algorithm implementation. Lines 6--15 of Algorithm~\ref{alg:learning} were executed within GradientTape context, what means that computation of $J(B_s,\mathrm{W},\mathrm{b})$ was recorded on a ``tape'', and finally in line 16 gradients $(\nabla_W J,\nabla_b J)$ were calculated via symbolic differentiation end evaluation of derivative expressions with respect to $\mathrm{W}$ and $\mathrm{b}$.

\subsubsection{Backpropagation}

Gradient computation with GradientTape turned out to be very useful as a prototyping aid, however, its time efficiency is limited. 
A significant slowdown can be observed for datests comprising a few hundreds attributes and for small batch sizes (if batches are small, gradient computations are more frequent).

Although all experiments described in Section~\ref{sec:experiments} were performed with a prototype based on GradientTape, we developed a more efficient  procedure for gradient calculation using internally the backpropagation mechanism.

\begin{figure}[ht!]
\centering
\includegraphics[width=1.0\linewidth]{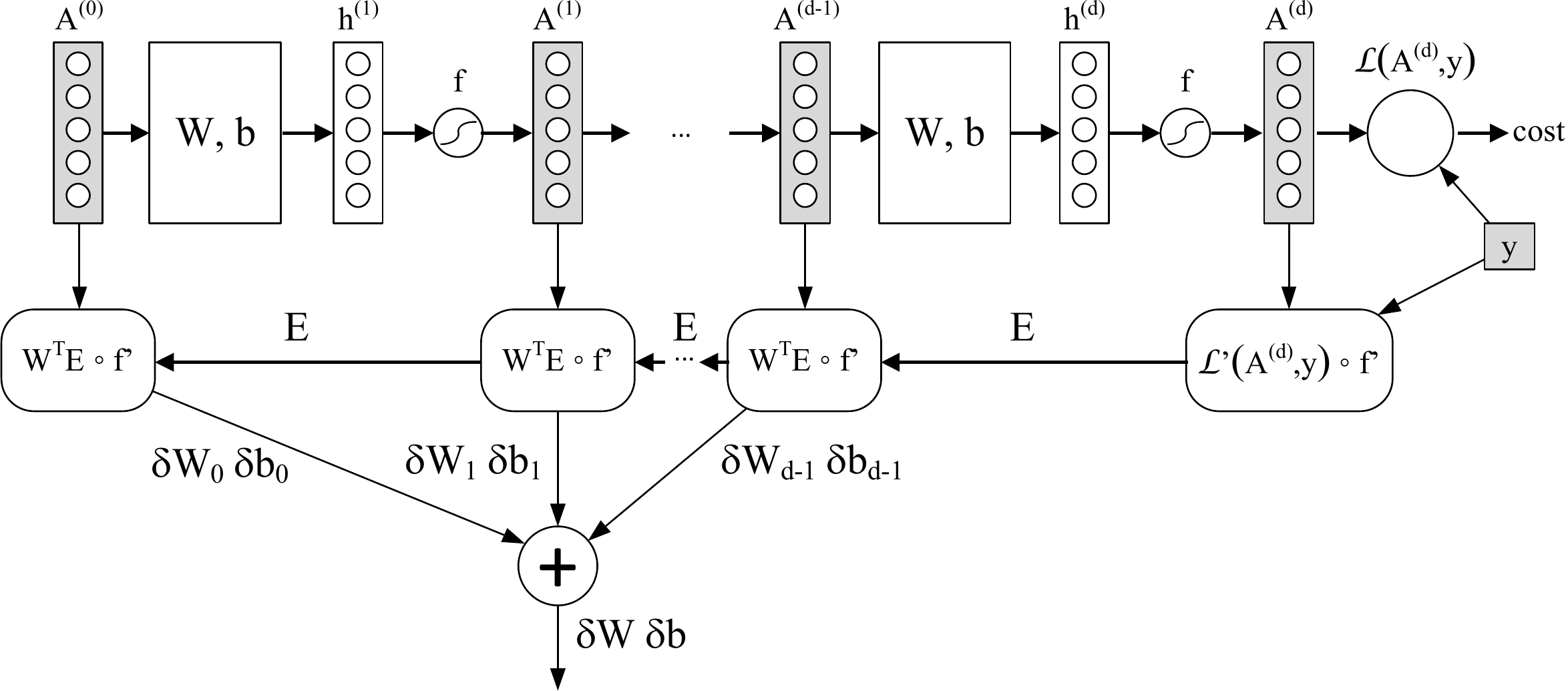}
\caption{Computation of gradient by FCM backpropagation}
\label{fig:fcm_backprop}
\end{figure}

As can be noticed in Fig.~\ref{fig:fcm_backprop}, Fuzzy Cognitive Map executed $d$ times can be unwound into a structure similar to feed forward neural networks with $d-1$ hidden layers. The main difference is that in the case of FCM all layers have equal sizes (i.e. numbers of neurons) and connection weights $\mathrm{W}$ and $\mathrm{b}$ are shared between layers. 

During backpropagation  in feed forward neural networks gradients, here marked as $(\delta \mathrm{W}_i, \delta \mathrm{b}_i)$, 
are computed with respect to disjoint sets of weights: $\{(\mathrm{W}_{i},\mathrm{b}_{i})\}_{i=0,d-1}$ and then are used to update them separately. 
For FCM, all gradients $\{(\delta \mathrm{W}_i, \delta \mathrm{b}_i)\}_{i=0,d-1}$ update shared weights $(\mathrm{W},\mathrm{b})$, hence, they should be added up. More formal derivation for a similar problem of RNN can be found in \cite{chen2016gentle}.

The procedure computing gradients of the cost function $J$ with respect to $W$ and $b$ is given in Algorithm~\ref{alg:back_prop}.  
We present the form close to actual implementation that uses matrix data representation. Its parameters include $r\times r$ weights matrix $W$, bias column vector $b$ of size $r$, FCM trajectory $\mathbf{A} = (\mathrm{A}^{(0)},\dots,\mathrm{A}^{(d)})$ and $y$, a row vector of size $m$; 
Trajectory elements $\mathrm{A}^{(i)}$ are $r\times m$ matrices, in which activation levels related to particular observations are arranged as columns. 
Dot product is marked with $\cdot$ and Hadamard product with $\circ$. 

Although derivative of activation function $f'(\mathrm{H}^{(i)})$, which takes $\mathrm{H}$ as parameter appears in lines 3 and 7, $\mathrm{H}$ values are not required, as $f'(\mathrm{H}^{(i)})$ can be computed from $\mathrm{A}^{(i)}$:
\begin{equation}
f'(\mathrm{H}^{(i)})=\lambda \cdot \mathrm{A}^{(i)}\circ([1]_{r\times m}-\mathrm{A}^{(i)})
\label{eq:sigmoid_derivative}
\end{equation}  

\begin{algorithm}[!ht]
\caption{Compute gradient of loss function $J$ with respect to weights $W$ and $b$}\label{alg:back_prop}
\begin{algorithmic}[1]
\Require
\Statex $\mathbf{A}=(A^{(0)},A^{(1)},\dots,A^{(d)})$ -- FCM trajectory, a sequence of $r \times m$ matrices, where $m$ is the number of observations
\Statex $d$ -- depth
\Statex $y$ -- ground truth labels (vector of $m$ elements)
\State $\mathrm{E} \gets \mathcal{L}'(\mathrm{A}^{(d)},y)$
\State $\mathrm{E} \gets \mathrm{E}\circ f'(\mathrm{H}^{(d)})$\Comment{Gradient of activation function  $f'$}
\For{$t = d-1$ \textbf{to} $0$}
	\State $\delta \mathrm{W}_t \gets \mathrm{E} \cdot \left(\mathrm{A}^{(i)}\right)^T$
	\State $\delta \mathrm{b}_t \gets \left[ \sum_{j=1}^m \mathrm{E}_{ij} \right] $ \Comment{Sum rows of $\mathrm{E}$}
	\State $E \gets (\mathrm{W}^T\cdot \mathrm{E}) \circ f'(\mathrm{H}^{(i)})$
\EndFor
\State $\delta \mathrm{W} \gets \sum_{t=0}^{d-1} \delta \mathrm{W}_t$ \Comment{Sum arrays $\delta \mathrm{W}$}
\State $\delta \mathrm{b} \gets \sum_{t=0}^{d-1} \delta \mathrm{b}_t$ \Comment{Sum vectors $\delta \mathrm{b}$}
\State \textbf{return} $\delta \mathrm{W}$, $\delta \mathrm{b}$ 
\end{algorithmic}
\end{algorithm}

Algorithm~\ref{alg:back_prop} was implemented using numpy Python package and its results were identical to those obtained with GradientTape. It was also about 20 times faster.

\section{Experiments}
\label{sec:experiments}

Experiments described in this section had two goals. The first was to evaluate performance of FCM based classifier on several datasets and to make comparison with such popular methods as Logistic Regression, Decision Tree, Random Forest, Na\"ive Bayes, SVM  and kNN.

The second goal was to verify the hypothesis that a trained FCM classifier is capable of transforming the original data by mapping them into a new feature space, referred here as $X^{(d-1)}$, in such a way that observations belonging to a given class become more condensed and easier to separate. 
Such supposition seemed to be true for the example presented in Section~\ref{sec:fcm-classifier:example}, c.f. 
Fig.~\ref{fig:ex_original_feature_space} and Fig.~\ref{fig:ex_transformed_feature_space}.

The above hypothesis can be tested in two ways: either by collecting metrics describing topological properties of $X^{(d-1)}$  or by using it as input to another classifier and comparing the classification scores for original and transformed feature spaces. Both approaches were realized: we collected clustering scores for  $X^{(d-1)}$, as well as performed classification in  $X^{(d-1)}$ using the same set of classifiers as in the first experiment.

\begin{figure}[ht!]
\includegraphics[width=1.0\linewidth]{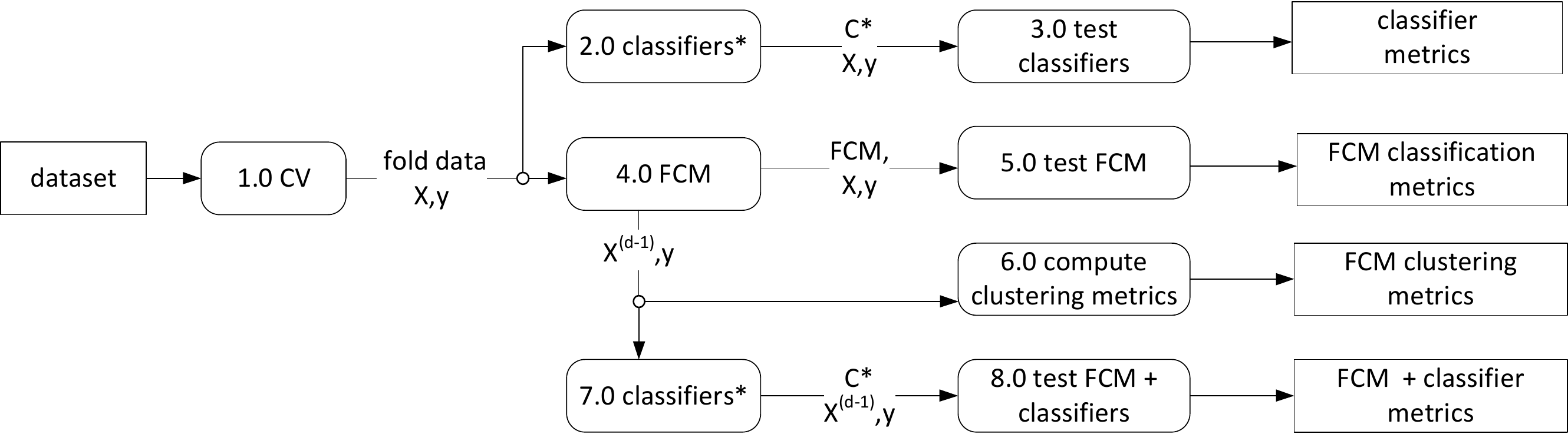}
\caption{Design of experiments}
\label{fig:experiments_design}
\end{figure}

Fig.~\ref{fig:experiments_design} shows data flow diagram of activities performed during experiments. First, an input dataset is split in \emph{1.0 CV} into $k=5$ folds. Each  operation that follows is carried out on the same fold data. Hence, classifiers and FCM are trained and tested on the same folds, what yields their classification metrics. Activity \emph{4.0 FCM} outputs transformed data (denoted here as $X^{(d-1)}$). They are used to compute clustering metrics in \emph{6.0} as well as to train and test classifiers  on transformed features in \emph{7.0} and \emph{8.0}. Actually, activities \emph{4.0},    
\emph{7.0} and \emph{8.0} can be considered a pipeline consisting of feature transformation and classification.

\subsection{Datasets}

Experiments were performed on 21 popular datasets that either can be found in UCI Machine Learning Repository \cite{Dua:2019} or loaded with built-in functions of scikit-learn and TensorFlow libraries.  They are summarized in Table~\ref{tab:datasets}. The selection of datasets was done based on their popularity, they are often used to test new algorithms.
Three variants of {\tt olivetti faces} dataset were obtained by scaling original $64 \times 64$ images to $8 \times 8$, $16 \times 16$ and $28 \times 28$ bitmaps. Dataset {\tt fashion 10000} is an abridged version of original dataset. Similarly, we used  shortened version of dataset {\tt digits}, which is provided  by scikit-learn library. 
Such limitations were introduced due to performance issues. As discussed earlier, we used a prototype implementation of FCM classifier, that uses internally relatively slow symbolic gradient computation.

\begin{table}[ht!]
\centering
{\scriptsize
\begin{tabular}{|c|l|c|c|c|}
\hline
No&Dataset&Attributes&Classes&Instances\\
\hline
1&iris&4&3&150\\
2&wine&13&3&178\\
3&brest cancer&30&2&569\\
4&glass&9&8&214\\
5&seeds&7&4&210\\
6&ionosphere&34&2&351\\
7&sonar&60&2&208\\
8&blood transfusion&4&2&748\\
9&vehicle&17&4&846\\
10&ecoli&6&8&336\\
11&yeast&7&10&1484\\
12&tic tac toe&8&2&958\\
13&heart&12&2&270\\
14&haberman&2&2&306\\
15&german credit&19&2&1000\\
16&diabets&7&2&768\\
17&olivetti faces 8&64&40&400\\
18&olivetti faces 16&256&40&400\\
19&olivetti faces 28&784&40&400\\
20&digits&64&10&1797\\
21&fashion10000&784&10&10000\\
\hline
\end{tabular}
}
\caption{Datasets used in experiments. }
\label{tab:datasets}
\end{table}

\subsection{Classifiers}
\label{sub:Classifiers}

During experiments we compared FCM classifier with selected methods available in the scikit-learn library.  The list of methods and configurations is given in Table~\ref{tab:classifiers}.


\begin{table}[ht!]
\centering
{\scriptsize
\begin{tabular}{|c|l|p{8cm}|}
\hline
1&mnb& Multinomial Na\"ive Bayes ($\alpha=0.01$)\\ 
2&knn3& k-NN ($k = 3$)\\
3&knn5& k-NN ($k = 5$)\\
4&svcrbf& SVC (Support Vector Machines) with RBF kernel and $\gamma$=scale\\
5&svclin& SVC with linear kernel\\
6&logreg& Logistic Regression\\
7&dtree& Decision Tree; the scikit-learn implementation supports only numeric attributes\\
8&rforest& Random Forest (10 trees)\\
9&gmc 1f&Gaussian Mixture ($n=1$, $\Sigma$=full)\\
10&gmc 3f&Gaussian Mixture ($n=3$, $\Sigma$=full)\\
11&gmc 3d&Gaussian Mixture ($n=3$, $\Sigma$=diag)\\
\hline
\end{tabular}
}
\caption{Classifers used in experiments.}
\label{tab:classifiers}
\end{table}

The last three configurations need an explanation. Basically, Gaussian Mixture (GM) is a clustering model that is learned with Expectation  Maximization algorithm. 
It provides probability density estimation of data in the form of a composite model comprising $n$ Gaussian components:
\begin{equation*}
M= \{M_i=(\mu_i,\Sigma_i)\colon i=1,\dots,n\},
\end{equation*}
The parameter  $\mu_i$ is a mean value of $i$-th component and $\Sigma_i$ its covariance matrix. 
Given a model $M$, the probability of observing $x$ can be estimated as:
\begin{equation*}
P(x|M) = \max\{P(x|M_i)\colon i=1,\dots,n\}
\end{equation*}

\noindent The procedure of turning it into a classifier consists in:
\begin{itemize}
\item Building GM models $M^{(j)}$, $j=1,\dots,k$ for each each subset of data $D_j=\{(x_i,y_i)\colon y_i=j\}$ with assigned class label $j$
\item Selecting the class label based on maximum posterior probability\\  
\begin{equation*}
\hat{y} = \mathrm{arg} \max_{j=1,k} \{P(x|M^{(j)})\cdot P(j)\}
\end{equation*}
\end{itemize}    

In this light, the  Gaussian Mixture Classifier (GMC) can be considered an extension to Gasussian Na\"ive Bayes allowing multiple components with probability density $P(x|M^{(j)}_i)$ scaled for each dimension (when $\Sigma$=diag) or scaled and rotated, when  $\Sigma$=full. The Gaussian Na\"ive Bayes classifier is equivalent to the configuration \emph{gmc 1d}: $n=1$, $\Sigma=diag$. 


\subsection{FCM parameters}

Behavior of FCM based classifier can be customized with a number of parameters. The most important are depth $d$ (number of FCM iterations) and $\lambda$ constant used in definition of activation function given by formula (\ref{eq:activation-fun}). Results of classifier learning depend also on number of epochs, batch size and learning rate $lr$. 

Parameters for particular datasets were selected  by performing a dozen trials, during which 80\% of data was used for training and 20\% for testing. The best configurations (typically, evaluation was based on F1 measure) were used in final experiments. The chosen configurations are listed in Table~\ref{tab:fcm_params}.

\begin{table}[ht!]
\centering
{\scriptsize
\begin{tabular}{|c|c|c|c|c|c|c|c|c|}
\hline
No&Dataset&Classifier&$d$&$\lambda$&epochs&$bs$&optimizer&$lr$\\
\hline
1&iris&FCMMC&4&3&3000&-1&rmsprop&0.0005\\
2&wine&FCMMC&4&1&3000&-1&rmsprop&0.001\\
3&brest cancer&FCMB&5&1&1000&-1&rmsprop&0.03\\
4&glass&FCMMC&2&1&3300&-1&rmsprop&0.02\\
5&seeds&FCMMC&2&1&3300&-1&rmsprop&0.08\\
6&ionosphere&FCMB&2&1&3300&-1&rmsprop&0.004\\
7&sonar&FCMB&2&1&500&-1&rmsprop&0.008\\
8&blood transfusion&FCMB&3&1&3300&-1&rmsprop&0.004\\
9&vehicle&FCMB&3&1&2000&-1&rmsprop&0.06\\
10&ecoli&FCMMC&2&2&5000&-1&adam&0.001\\
11&yeast&FCMMC&3&2.8&5000&-1&rmsprop&0.032\\
12&tic tac toe&FCMMC&3&2&5000&-1&adam&0.001\\
13&heart&FCMMC&3&2&5000&-1&adam&0.001\\
14&haberman&FCMMC&3&2&5000&-1&adam&0.001\\
15&german credit&FCMMC&2&1&5000&-1&adam&0.001\\
16&diabets&FCMMC&3&1&3000&-1&adam&0.001\\
17&olivetti faces 8&FCMMC&3&1&5000&-1&rmsprop&0.00045\\
18&olivetti faces 16&FCMMC&3&1&5000&-1&rmsprop&0.00045\\
19&olivetti faces 28&FCMMC&3&1&4000&-1&rmsprop&0.006\\
20&digits&FCMMC&3&0.5&120&20&rmsprop&0.01\\
21&fashion10000&FCMMC&3&1&600&1000&rmsprop&0.005\\
\hline
\end{tabular}
}
\caption{FCM parameters: $d$ -- depth; $\lambda$ constant used in definition of activation function given by formula (\ref{eq:activation-fun}); $bs$ -- batch size, value -1 is put, if a whole dataset is processed as a single batch, $lr$ - learning rate }
\label{tab:fcm_params}
\end{table}

\section{Results}
\label{sec:results}

This section reports results of three experiments: 

\begin{enumerate}
\item Classification. In the first experiments we compared performance of FCM based classifier and several classifiers discussed in Section~\ref{sub:Classifiers}
\item Feature transformation. In the second experiment we calculated clustering scores to compare compactness and separability of two feature spaces: original -- $X^{(0)}$ and transformed by FCM -- $X^{(d-1)}$  
\item Preprocessing. In the third experiment we compared performance of FCM and classifiers, which were trained on the transformed feature space $X^{(d-1)}$   
\end{enumerate}

\subsection{Classification results}

Classifier performance was evaluated using two popular metrics: accuracy and F1 macro score \cite{murphy2012machine}. 
%
%
%
%


Table~\ref{tab:accuracy} and Table~\ref{tab:f1_score} give mean values of accuracy and F1 score obtained by 5-fold cross validation for datasets used in experiments. The results show that FCM based classifier is competitive to the other, it yielded the best mean scores 4 times for accuracy, and 3 times for F1, although in certain cases, e.g. for \emph{yeast} dataset it failed.

\begin{table}[ht!]
\centering
{\tiny

\begin{tabular}{|l|c|c|c|c|c|c|c|c|c|c|c|c|c|}
\hline
&fcm&mnb&gnb&knn3&knn5&svcrbf&svclin&logreg&dtree&rforest&gmc 1f&gmc 3f&gmc 3d\\
\hline
iris&$0.97$&$0.77$&$0.95$&$0.95$&$0.96$&$0.97$&$0.97$&$0.84$&$0.96$&$0.97$&$\mathbf{ 0.98}$&$0.97$&$0.96$\\
wine&$0.97$&$0.93$&$0.96$&$0.95$&$0.96$&$0.97$&$0.97$&$0.97$&$0.85$&$\mathbf{ 0.98}$&$0.96$&$0.86$&$0.96$\\
brest cancer&$0.96$&$0.84$&$0.93$&$0.96$&$0.97$&$0.97$&$\mathbf{ 0.98}$&$0.96$&$0.92$&$0.96$&$0.96$&$0.95$&$0.93$\\
glass&$0.58$&$0.43$&$0.32$&$0.64$&$0.64$&$0.57$&$0.54$&$0.48$&$0.62$&$\mathbf{ 0.66}$&$0.50$&$0.52$&$0.46$\\
seeds&$\mathbf{ 0.94}$&$0.89$&$0.89$&$0.91$&$0.92$&$0.93$&$0.93$&$0.91$&$0.90$&$0.90$&$0.93$&$0.90$&$0.89$\\
ionosphere&$0.90$&$0.66$&$0.87$&$0.83$&$0.83$&$0.90$&$0.87$&$0.85$&$0.86$&$\mathbf{ 0.91}$&$0.89$&$0.81$&$0.88$\\
sonar&$\mathbf{ 0.68}$&$0.66$&$0.64$&$0.59$&$0.54$&$0.61$&$0.63$&$0.65$&$0.61$&$0.67$&$0.59$&$0.55$&$0.56$\\
blood transfusion&$0.75$&$0.76$&$0.75$&$0.63$&$0.65$&$0.76$&$0.76$&$\mathbf{ 0.77}$&$0.62$&$0.66$&$0.65$&$0.63$&$0.61$\\
vehicle&$0.75$&$0.53$&$0.45$&$0.70$&$0.69$&$0.71$&$0.71$&$0.67$&$0.72$&$0.74$&$\mathbf{ 0.83}$&$0.80$&$0.67$\\
ecoli&$0.78$&$0.44$&$0.72$&$0.79$&$0.78$&$\mathbf{ 0.80}$&$0.80$&$0.75$&$0.72$&$0.79$&$0.74$&$0.74$&$0.76$\\
yeast&$0.45$&$0.33$&$0.14$&$0.50$&$0.53$&$\mathbf{ 0.53}$&$0.53$&$0.53$&$0.43$&$0.52$&$0.19$&$0.41$&$0.38$\\
tic tac toe&$0.75$&$0.67$&$0.67$&$0.76$&$\mathbf{ 0.81}$&$0.73$&$0.65$&$0.67$&$0.76$&$0.75$&$0.56$&$0.70$&$0.58$\\
heart&$0.81$&$0.82$&$0.84$&$0.80$&$0.81$&$0.83$&$\mathbf{ 0.85}$&$0.83$&$0.72$&$0.78$&$0.82$&$0.71$&$0.75$\\
haberman&$0.72$&$0.74$&$0.74$&$0.69$&$0.68$&$0.73$&$0.74$&$0.74$&$0.72$&$0.73$&$\mathbf{ 0.76}$&$0.64$&$0.64$\\
german credit&$0.70$&$0.70$&$0.70$&$0.69$&$0.71$&$0.71$&$0.69$&$\mathbf{ 0.71}$&$0.65$&$0.70$&$0.64$&$0.59$&$0.57$\\
diabets&$\mathbf{ 0.78}$&$0.65$&$0.75$&$0.74$&$0.74$&$0.77$&$0.77$&$0.76$&$0.70$&$0.74$&$0.74$&$0.66$&$0.66$\\
olivetti faces 8&$0.86$&$0.76$&$0.82$&$0.85$&$0.79$&$0.70$&$0.95$&$0.92$&$0.48$&$0.77$&$\mathbf{ 0.97}$&$0.97$&$0.82$\\
olivetti faces 16&$0.91$&$0.83$&$0.85$&$0.86$&$0.81$&$0.81$&$\mathbf{ 0.97}$&$0.97$&$0.60$&$0.78$&$0.97$&$0.97$&$0.85$\\
olivetti faces 28&$0.73$&$0.86$&$0.88$&$0.90$&$0.83$&$0.85$&$\mathbf{ 0.98}$&$0.98$&$0.64$&$0.81$&$0.97$&$0.97$&$0.88$\\
digits&$0.94$&$0.87$&$0.80$&$\mathbf{ 0.97}$&$0.96$&$0.95$&$0.95$&$0.93$&$0.78$&$0.90$&$0.92$&$0.73$&$0.86$\\
fashion10000&$\mathbf{ 0.85}$&$0.67$&$0.52$&$0.82$&$0.82$&$0.84$&$0.83$&$0.83$&$0.75$&$0.82$&$0.67$&$0.65$&$0.62$\\
\hline
\end{tabular}

}
\caption{Mean accuracy obtained during 5-fold CV. Best values are marked with boldface.}
\label{tab:accuracy}
\end{table}

\begin{table}[ht!]
\centering
{\tiny

\begin{tabular}{|l|c|c|c|c|c|c|c|c|c|c|c|c|c|}
\hline
&fcm&mnb&gnb&knn3&knn5&svcrbf&svclin&logreg&dtree&rforest&gmc 1f&gmc 3f&gmc 3d\\
\hline
iris&$0.97$&$0.77$&$0.95$&$0.95$&$0.96$&$0.97$&$0.97$&$0.83$&$0.96$&$0.97$&$\mathbf{ 0.98}$&$0.97$&$0.96$\\
wine&$0.97$&$0.93$&$0.96$&$0.95$&$0.96$&$0.97$&$0.97$&$0.97$&$0.85$&$\mathbf{ 0.98}$&$0.96$&$0.86$&$0.96$\\
brest cancer&$0.96$&$0.81$&$0.92$&$0.96$&$0.96$&$0.96$&$\mathbf{ 0.97}$&$0.96$&$0.92$&$0.96$&$0.96$&$0.95$&$0.93$\\
glass&$0.50$&$0.24$&$0.34$&$0.55$&$0.47$&$0.35$&$0.33$&$0.29$&$0.54$&$\mathbf{ 0.59}$&$0.43$&$0.48$&$0.41$\\
seeds&$\mathbf{ 0.94}$&$0.89$&$0.89$&$0.91$&$0.92$&$0.93$&$0.93$&$0.91$&$0.90$&$0.90$&$0.93$&$0.90$&$0.89$\\
ionosphere&$0.88$&$0.44$&$0.85$&$0.79$&$0.79$&$0.88$&$0.84$&$0.81$&$0.85$&$\mathbf{ 0.89}$&$0.89$&$0.80$&$0.88$\\
sonar&$0.67$&$0.65$&$0.64$&$0.58$&$0.54$&$0.61$&$0.62$&$0.65$&$0.60$&$\mathbf{ 0.67}$&$0.56$&$0.48$&$0.55$\\
blood transfusion&$0.54$&$0.43$&$0.55$&$0.47$&$0.47$&$0.44$&$0.43$&$0.49$&$0.48$&$0.51$&$\mathbf{ 0.59}$&$0.57$&$0.53$\\
vehicle&$0.75$&$0.52$&$0.43$&$0.70$&$0.68$&$0.69$&$0.70$&$0.65$&$0.72$&$0.73$&$\mathbf{ 0.83}$&$0.80$&$0.67$\\
ecoli&$0.51$&$0.22$&$0.46$&$0.65$&$0.62$&$0.59$&$0.59$&$0.49$&$0.46$&$\mathbf{ 0.65}$&$0.49$&$0.60$&$0.59$\\
yeast&$0.23$&$0.12$&$0.24$&$0.44$&$\mathbf{ 0.47}$&$0.34$&$0.34$&$0.29$&$0.28$&$0.37$&$0.21$&$0.36$&$0.36$\\
tic tac toe&$0.73$&$0.51$&$0.62$&$0.73$&$\mathbf{ 0.79}$&$0.68$&$0.58$&$0.61$&$0.73$&$0.72$&$0.52$&$0.68$&$0.57$\\
heart&$0.81$&$0.82$&$0.84$&$0.79$&$0.81$&$0.83$&$\mathbf{ 0.85}$&$0.83$&$0.72$&$0.78$&$0.82$&$0.71$&$0.75$\\
haberman&$0.50$&$0.42$&$0.56$&$0.54$&$0.52$&$0.42$&$0.42$&$0.45$&$0.56$&$0.58$&$\mathbf{ 0.63}$&$0.58$&$0.60$\\
german credit&$0.60$&$0.41$&$\mathbf{ 0.64}$&$0.59$&$0.60$&$0.46$&$0.43$&$0.56$&$0.60$&$0.57$&$0.61$&$0.53$&$0.52$\\
diabets&$\mathbf{ 0.75}$&$0.39$&$0.72$&$0.71$&$0.70$&$0.73$&$0.73$&$0.71$&$0.67$&$0.70$&$0.71$&$0.63$&$0.66$\\
olivetti faces 8&$0.83$&$0.75$&$0.80$&$0.84$&$0.78$&$0.66$&$0.94$&$0.91$&$0.45$&$0.75$&$\mathbf{ 0.97}$&$0.97$&$0.80$\\
olivetti faces 16&$0.89$&$0.82$&$0.84$&$0.85$&$0.80$&$0.79$&$\mathbf{ 0.97}$&$0.97$&$0.56$&$0.76$&$0.97$&$0.97$&$0.84$\\
olivetti faces 28&$0.70$&$0.86$&$0.87$&$0.89$&$0.82$&$0.84$&$\mathbf{ 0.98}$&$0.97$&$0.62$&$0.80$&$0.97$&$0.97$&$0.87$\\
digits&$0.94$&$0.87$&$0.80$&$\mathbf{ 0.97}$&$0.96$&$0.95$&$0.95$&$0.93$&$0.78$&$0.90$&$0.92$&$0.73$&$0.86$\\
fashion10000&$\mathbf{ 0.84}$&$0.64$&$0.47$&$0.81$&$0.81$&$0.83$&$0.83$&$0.83$&$0.75$&$0.82$&$0.64$&$0.63$&$0.61$\\
\hline
\end{tabular}

}
\caption{Mean F1 scores obtained during 5-fold CV. Best values are marked with boldface.}
\label{tab:f1_score}
\end{table}

Since all experiments were performed on the same folds (c.f. Fig.~\ref{fig:experiments_design}), classifiers performance can be also compared taking into account results for particular fold data. During 5-fold CV each of 21 datasets was divided into 5 folds, what gives 105 test samples for each classifier.
Analysis of larger samples provides better insight into variability of results.

\begin{figure}[h]
{\scriptsize
\begin{tabular}{c}
\includegraphics[width=0.9\linewidth]{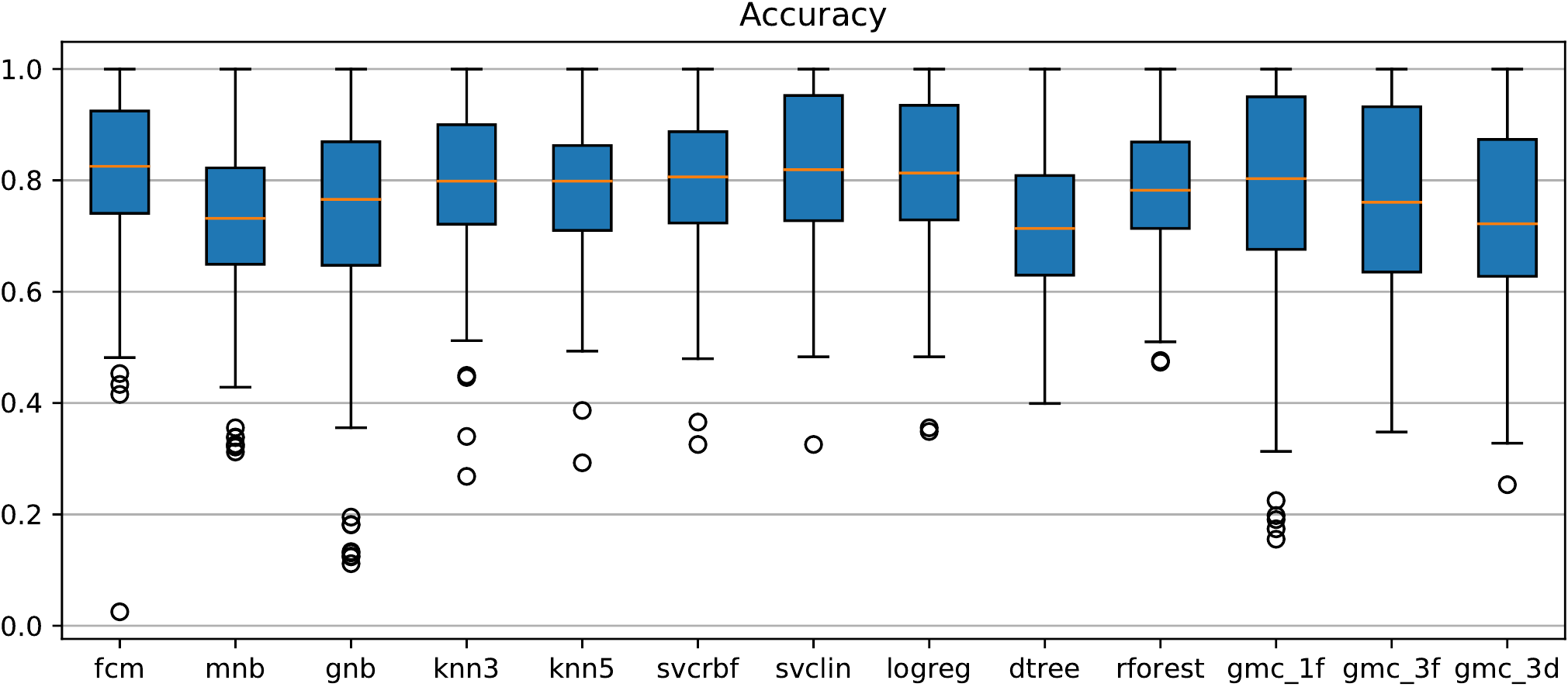}\\
(a)\\
\includegraphics[width=0.9\linewidth]{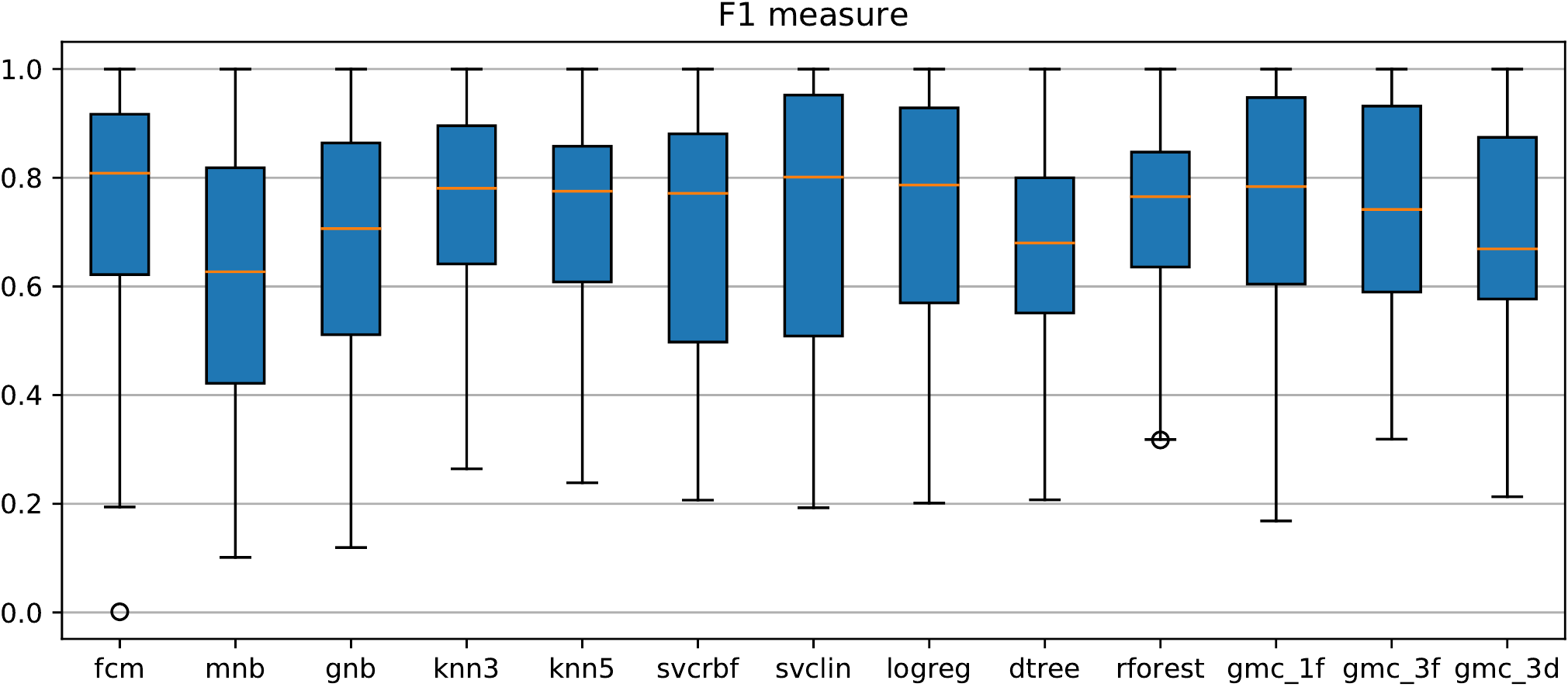}\\
(b)
\end{tabular}
}
\caption{Comparison of (a) accuracy and (b) F1 macro scores computed on folds. Outliers marked with circles. }
\label{fig:boxplots}
\end{figure}

Fig.~\ref{fig:boxplots} shows boxplots  of accuracy and F1 score computed on folds. 
Surprisingly, the medians of accuracy and F1 calculated over folds was slightly higher in the case of FCM classifier, than the other. 
On the other hand, Q3 quartile was apparently higher for \emph{svclin} (SVM with linear kernel), \emph{logreg} (Logistic Regression) and \emph{gmc\_1f} (Gaussian Mixture with full covariance matrix). Circles  on a boxplots are used to mark outliers. In the accuracy plots Fig.~\ref{fig:boxplots}a they are present for the majority of classifiers. For F1 plots, the only outlier was indicated for FCM. Analysis of results revealed that for just one fold of \emph{olliveti faces 28} accuracy and F1 were close to 0, whereas for the other folds, both scores were around 0.9.

A typical procedure for statistical evaluation of classifiers performance on multiple datasets  \cite{japkowicz2011evaluating} comprises two steps: first omnibus tests are carried out to reject the hypothesis that differences between classifier performance for various datasets are random. Next, post-hoc tests are performed to indicate groups of classifiers, such that the performance within a group cannot be discerned.   

We performed  two omnibus tests: repeated-measures one-way ANOVA and Friedman test  on a group of classifiers and datasets. ANOVA is based on median performance obtained by a classifier for a group of datasets. For Friedman test the analysis takes into account ranks, which are assigned to classifiers by comparing their performance measures for each dataset. 
In both cases the null hypothesis is that the difference of classifiers performance observed within a group of datasets is statistically insignificant.

The tests for accuracy and F1 scores were performed both on 21 datasets (i.e. mean values obtained by 5-fold CV) and on 105 folds treated as separate items. Their results are summarized in Table~\ref{tab:omnibus_classifiers}. 
It can be observed that ANOVA was not capable of rejecting the null hypothesis for measures computed for the datasets ($p$-value for accuracy was equal 0.19 and 0.41 for F1). This can be attributed to limited sample size. On the other hand, the null hypothesis was rejected with ANOVA for folds. In all cases hypotheses that classifiers performance is the same were rejected with very high significance levels with Friedman test.

As the rank based Friedman test showed greater sensitivity, we performed post hoc Nemenyi test. The test computes average ranks  for each classifier and  analyzes differences between them. If the difference between average ranks  of two classifiers $C_i$ and $C_j$  is greater than a statistics called \emph{critical difference} (CD), it can be stated that the performance of $C_i$ and $C_j$ differ significantly.  

\begin{figure}[ht!]
{\scriptsize
\begin{tabular}{c}
\includegraphics[width=0.9\linewidth]{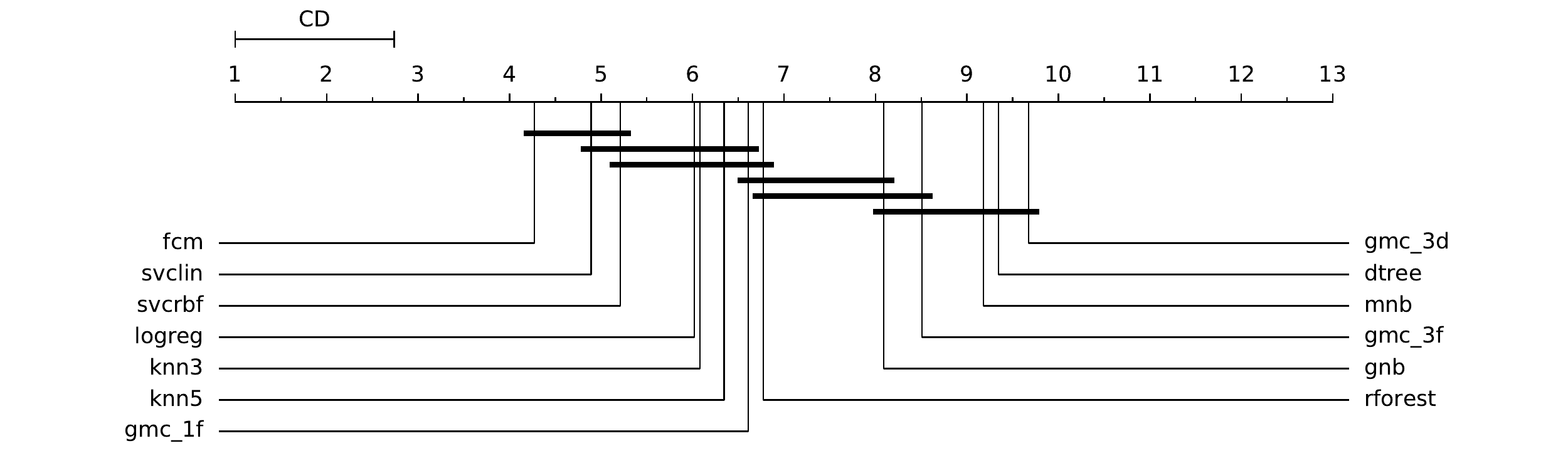}\\
(a) Accuracy\\
\includegraphics[width=0.9\linewidth]{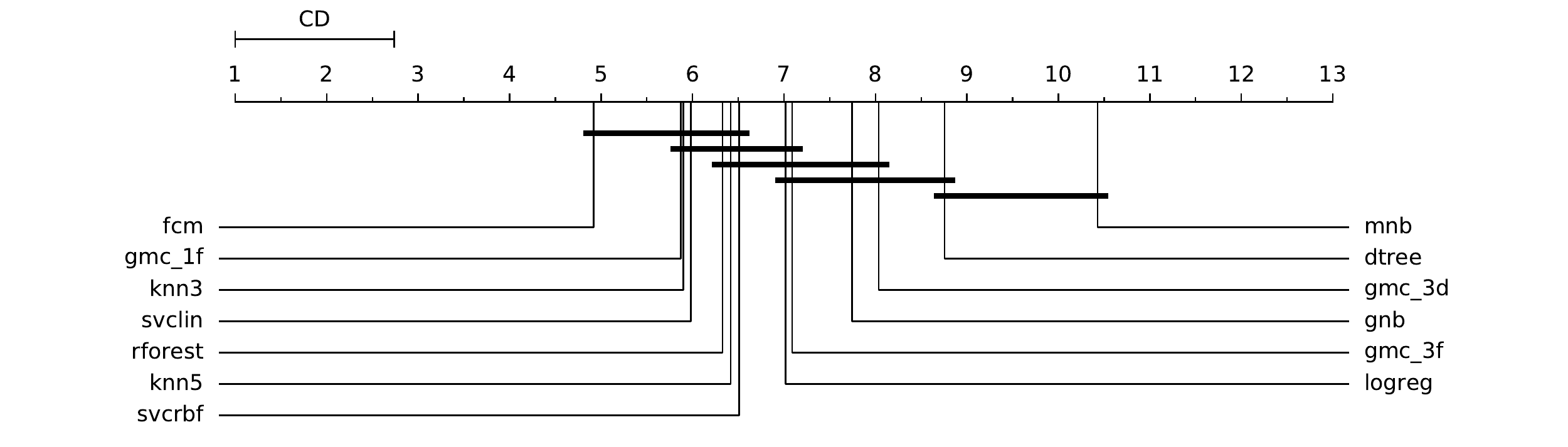}\\
(b) F1 macro
\end{tabular}
}
\caption{Results of Nemenyi post hoc tests for (a) accuracy (b) F1 macro score computed on folds. Classifiers are ordered according to their average ranks (\emph{smaller}=\emph{better}). Bold lines link classifiers that could not be discerned according to critical difference (CD) computed during the test.}
\label{fig:ranks_nemenyi}
\end{figure}

Results of Nemenyi tests for accuracy and F1 computed over folds are shown in Fig.~\ref{fig:ranks_nemenyi}. 
It can be seen that average ranks were the best for FCM classifier. As regards accuracy, FCM formed a common group with two variants of SVM. 
For F1 score, its performance cannot be discerned at assumed level of certainty from a larger group of classifiers comprising Gaussian Mixture, kNN, SVM and Random Forest.

\begin{table}[ht!]
\centering
{\scriptsize

\begin{tabular}{|l|c|c|c|c|}
\hline
&\multicolumn{2}{c|}{Accuracy}&\multicolumn{2}{c|}{F1 macro}\\
\cline{2-3}\cline{3-5}
&datasets&folds&datasets&folds\\
\hline

\hline
ANOVA $F$&1.36&\textbf{6.07}&1.04&\textbf{4.88}\\
ANOVA $p$-value&0.19&\textbf{1.8e-10}&0.41&\textbf{6.1e-08}\\
\hline
Friedman $\chi^2_F$&\textbf{76.21}&\textbf{258.11}&\textbf{54.03}&\textbf{184.68}\\
Friedman $p$-value&\textbf{2.2e-11}&\textbf{2.8e-48}&\textbf{2.7e-07}&\textbf{4.7e-33}\\
\hline
\end{tabular}

}
\caption{Results of omnibus test for accuracy and F1 macro score performed on average results for datasets and individual results for folds. Test in which the null hypothesis $H_0$ was rejected are marked with boldface.}
\label{tab:omnibus_classifiers}
\end{table}


%
%
%
%
%
%
%
%
%

\subsection{Feature transformation}

We designed experiments to verify the hypothesis that a trained FCM classifier transforms the original data,  so that the final classification step, which is equivalent to Logistic Regression, becomes easier. We supposed that topological properties of data change: observations belonging to a given class become more condensed and  separable. This two properties correspond to  well known internal clustering evaluation criteria: \emph{compactness} and \emph{separation} \cite{liu2010understanding}.

For a given set of observations $X=\{x_i\}_{i=1,m}$, clustering can be defined as a function $Z\colon \{1,\dots,m\} \to L$ that assigns cluster labels from $L\subset \mathbb{N}$ to observation indexes. 

Internal clustering validation methods use real valued scores  $s(X,Z)$ to compare two possible clusterings $Z_1$ and $Z_2$ for a given dataset $X$. Clustering $Z_1$ is considered better then $Z_2$ for  $X$ if  $s(X,Z_1) \succ s(X,Z_2)$. Depending on the score type,  $<$ or $>$ should be plugged in the place of  $\succ$.

The clustering scores are typically used to answer the question: {\it is clustering $Z_1$ better than $Z_2$ for a particular dataset $X$}. We address here another problem: {\it given mapping $Z$ resulting from known ground-truth labels, is the transformed dataset better than the original}.

For an input dataset $D=\{(x_i,y_i)\}_{i=1,m}$, which comprises observations $x_i$ and their ground truth labels $y_i$, the function $Z$ can be defined as $Z=\{(i,y_i)\}_{i=1,m}$. 
Keeping $Z$ fixed,  we compare scores for $X^{(0)}$ and $X^{(d-1)}$.  From the fact that $s(X^{(d-1},Z)\succ s(X^{(0)},Z)$, we may state that the transformed dataset $X^{(d-1)}$ exhibits the desired properties to a greater extent.

We decided to apply three internal clustering scores Davies-Bouldin \cite{davies1979cluster}, Silhouette Coefficient \cite{liu2010understanding} and Calinski-Harabasz index \cite{calinski1974dendrite}.

\paragraph{Results}
Collected clustering metrics are given in Table~\ref{tab:clustering_scores}. The table comprises values averaged over 5 folds for training and test datasets. To confirm the hypothesis, that a trained FCM classifier is capable of improving class compactness and separability,  we applied  \emph{majority voting} among three scores. 

For 15 out of 21 datasets the assumed effect was confirmed unanimously. 
For three datasets: (\emph{german credit}, \emph{blood transfusion} and \emph{ecoli}) there were 2:1 votes in favor of the hypothesis. For two datasets: \emph{glass} and \emph{yeast} the voting was opposite. Analyzing accuracy and F1 scores in Table~\ref{tab:accuracy} and Table~\ref{tab:f1_score} we may notice that for these particular two datasets the performance of FCM classifier was poor. Moreover, average accuracy and F1 scores for training fold data were close to testing results.        
This fact can be also considered as support for our hypothesis: we may state that  FCM classifier was not properly trained (or not trainable due to data noise or innate model bias) and in consequence the expected data transformation did not occur.

\begin{table}[ht!]
\centering
{\scriptsize

\begin{tabular}{||l||c|c||c|c||c|c||}
\hline
Dataset&DB orig&DB transf&SLH orig&SLH transf&CH orig&CH transf\\
\hline
iris&\shortstack{ \\ 0.87 \\ \textbf{ 0.84 } }&\shortstack{ \\ 0.36 \\ \textbf{ 0.37 } }&\shortstack{ \\ 0.46 \\ \textbf{ 0.46 } }&\shortstack{ \\ 0.77 \\ \textbf{ 0.76 } }&\shortstack{ \\ 252.51 \\ \textbf{ 68.64 } }&\shortstack{ \\ 669.07 \\ \textbf{ 194.82 } }\\
\hline
wine&\shortstack{ \\ 1.31 \\ \textbf{ 1.20 } }&\shortstack{ \\ 0.24 \\ \textbf{ 0.34 } }&\shortstack{ \\ 0.29 \\ \textbf{ 0.32 } }&\shortstack{ \\ 0.84 \\ \textbf{ 0.76 } }&\shortstack{ \\ 64.76 \\ \textbf{ 19.66 } }&\shortstack{ \\ 1246.62 \\ \textbf{ 150.84 } }\\
\hline
brest cancer&\shortstack{ \\ 1.22 \\ \textbf{ 1.20 } }&\shortstack{ \\ 0.28 \\ \textbf{ 0.34 } }&\shortstack{ \\ 0.34 \\ \textbf{ 0.34 } }&\shortstack{ \\ 0.80 \\ \textbf{ 0.75 } }&\shortstack{ \\ 251.74 \\ \textbf{ 66.05 } }&\shortstack{ \\ 3820.31 \\ \textbf{ 513.49 } }\\
\hline
glass&\shortstack{ \\ 4.09 \\ \textbf{ 2.16 } }&\shortstack{ \\ 3.93 \\ \textbf{ 2.46 } $\bullet$}&\shortstack{ \\ -0.05 \\ \textbf{ 0.00 } }&\shortstack{ \\ 0.04 \\ \textbf{ -0.00 } $\bullet$ }&\shortstack{ \\ 20.81 \\ \textbf{ 11.38 } }&\shortstack{ \\ 28.17 \\ \textbf{ 7.91 } $\bullet$ }\\
\hline
seeds&\shortstack{ \\ 0.93 \\ \textbf{ 0.88 } }&\shortstack{ \\ 0.73 \\ \textbf{ 0.74 } }&\shortstack{ \\ 0.38 \\ \textbf{ 0.42 } }&\shortstack{ \\ 0.54 \\ \textbf{ 0.54 } }&\shortstack{ \\ 219.51 \\ \textbf{ 67.24 } }&\shortstack{ \\ 251.18 \\ \textbf{ 74.09 } }\\
\hline
ionosphere&\shortstack{ \\ 3.85 \\ \textbf{ 3.38 } }&\shortstack{ \\ 2.10 \\ \textbf{ 2.49 } }&\shortstack{ \\ 0.16 \\ \textbf{ 0.17 } }&\shortstack{ \\ 0.29 \\ \textbf{ 0.24 } }&\shortstack{ \\ 16.15 \\ \textbf{ 5.53 } }&\shortstack{ \\ 56.94 \\ \textbf{ 11.10 } }\\
\hline
sonar&\shortstack{ \\ 4.80 \\ \textbf{ 2.68 } }&\shortstack{ \\ 2.26 \\ \textbf{ 2.52 } }&\shortstack{ \\ 0.04 \\ \textbf{ 0.12 } }&\shortstack{ \\ 0.16 \\ \textbf{ 0.14 } }&\shortstack{ \\ 7.03 \\ \textbf{ 5.90 } }&\shortstack{ \\ 31.96 \\ \textbf{ 7.68 } }\\
\hline
blood transfusion&\shortstack{ \\ 4.62 \\ \textbf{ 3.55 } }&\shortstack{ \\ 2.97 \\ \textbf{ 4.10 $\bullet$} }&\shortstack{ \\ 0.05 \\ \textbf{ 0.07 } }&\shortstack{ \\ 0.12 \\ \textbf{ 0.12 } }&\shortstack{ \\ 14.55 \\ \textbf{ 6.26 } }&\shortstack{ \\ 59.57 \\ \textbf{ 12.58 } }\\
\hline
vehicle&\shortstack{ \\ 10.11 \\ \textbf{ 6.69 } }&\shortstack{ \\ 5.13 \\ \textbf{ 5.46 } }&\shortstack{ \\ -0.02 \\ \textbf{ -0.03 } }&\shortstack{ \\ 0.12 \\ \textbf{ 0.10 } }&\shortstack{ \\ 34.39 \\ \textbf{ 9.20 } }&\shortstack{ \\ 123.04 \\ \textbf{ 28.21 } }\\
\hline
ecoli&\shortstack{ \\ 2.68 \\ \textbf{ 1.68 } }&\shortstack{ \\ 2.37 \\ \textbf{ 1.74 $\bullet$ } }&\shortstack{ \\ 0.14 \\ \textbf{ 0.15 } }&\shortstack{ \\ 0.20 \\ \textbf{ 0.20 } }&\shortstack{ \\ 73.05 \\ \textbf{ 25.67 } }&\shortstack{ \\ 97.75 \\ \textbf{ 29.69 } }\\
\hline
yeast&\shortstack{ \\ 2.91 \\ \textbf{ 3.01 } }&\shortstack{ \\ 3.62 \\ \textbf{ 4.02 $\bullet$} }&\shortstack{ \\ -0.01 \\ \textbf{ -0.04 } }&\shortstack{ \\ -0.04 $\bullet$ \\ \textbf{ -0.06 $\bullet$ } }&\shortstack{ \\ 52.55 \\ \textbf{ 14.66 } }&\shortstack{ \\ 70.61 \\ \textbf{ 17.91 } }\\
\hline
tic tac toe&\shortstack{ \\ 5.67 \\ \textbf{ 4.28 } }&\shortstack{ \\ 1.90 \\ \textbf{ 2.79 } }&\shortstack{ \\ 0.04 \\ \textbf{ 0.07 } }&\shortstack{ \\ 0.23 \\ \textbf{ 0.12 } }&\shortstack{ \\ 21.59 \\ \textbf{ 10.31 } }&\shortstack{ \\ 200.93 \\ \textbf{ 21.95 } }\\
\hline
heart&\shortstack{ \\ 2.48 \\ \textbf{ 2.38 } }&\shortstack{ \\ 0.87 \\ \textbf{ 1.23 } }&\shortstack{ \\ 0.13 \\ \textbf{ 0.13 } }&\shortstack{ \\ 0.45 \\ \textbf{ 0.31 } }&\shortstack{ \\ 32.94 \\ \textbf{ 8.87 } }&\shortstack{ \\ 218.97 \\ \textbf{ 28.94 } }\\
\hline
haberman&\shortstack{ \\ 6.37 \\ \textbf{ 4.81 } }&\shortstack{ \\ 5.29 \\ \textbf{ 4.50 } }&\shortstack{ \\ 0.06 \\ \textbf{ 0.06 } }&\shortstack{ \\ 0.07 \\ \textbf{ 0.07 } }&\shortstack{ \\ 3.96 \\ \textbf{ 2.13 } }&\shortstack{ \\ 7.71 \\ \textbf{ 2.30 } }\\
\hline
german credit&\shortstack{ \\ 10.21 \\ \textbf{ 8.13 } }&\shortstack{ \\ 5.52 \\ \textbf{ 6.82 } }&\shortstack{ \\ 0.01 \\ \textbf{ 0.01 } }&\shortstack{ \\ 0.01 \\ \textbf{ 0.0} $\bullet$ }&\shortstack{ \\ 6.30 \\ \textbf{ 2.50 } }&\shortstack{ \\ 20.32 \\ \textbf{ 3.46 } }\\
\hline
diabets&\shortstack{ \\ 3.54 \\ \textbf{ 3.45 } }&\shortstack{ \\ 1.56 \\ \textbf{ 1.62 } }&\shortstack{ \\ 0.09 \\ \textbf{ 0.09 } }&\shortstack{ \\ 0.22 \\ \textbf{ 0.21 } }&\shortstack{ \\ 39.65 \\ \textbf{ 10.47 } }&\shortstack{ \\ 189.88 \\ \textbf{ 46.55 } }\\
\hline
olivetti faces 8&\shortstack{ \\ 1.93 \\ \textbf{ 1.47 } }&\shortstack{ \\ 1.42 \\ \textbf{ 1.12 } }&\shortstack{ \\ 0.10 \\ \textbf{ 0.06 } }&\shortstack{ \\ 0.25 \\ \textbf{ 0.20 } }&\shortstack{ \\ 13.62 \\ \textbf{ 4.03 } }&\shortstack{ \\ 18.14 \\ \textbf{ 4.78 } }\\
\hline
olivetti faces 16&\shortstack{ \\ 1.82 \\ \textbf{ 1.28 } }&\shortstack{ \\ 1.34 \\ \textbf{ 1.06 } }&\shortstack{ \\ 0.13 \\ \textbf{ 0.12 } }&\shortstack{ \\ 0.27 \\ \textbf{ 0.22 } }&\shortstack{ \\ 12.03 \\ \textbf{ 4.14 } }&\shortstack{ \\ 15.47 \\ \textbf{ 4.24 } }\\
\hline
olivetti faces 28&\shortstack{ \\ 1.86 \\ \textbf{ 1.32 } }&\shortstack{ \\ 1.64 \\ \textbf{ 1.17 } }&\shortstack{ \\ 0.12 \\ \textbf{ 0.09 } }&\shortstack{ \\ 0.21 \\ \textbf{ 0.14 } }&\shortstack{ \\ 10.66 \\ \textbf{ 3.34 } }&\shortstack{ \\ 11.21 \\ \textbf{ 3.12 } }\\
\hline
digits&\shortstack{ \\ 2.13 \\ \textbf{ 1.86 } }&\shortstack{ \\ 0.73 \\ \textbf{ 0.92 } }&\shortstack{ \\ 0.16 \\ \textbf{ 0.20 } }&\shortstack{ \\ 0.55 \\ \textbf{ 0.47 } }&\shortstack{ \\ 116.14 \\ \textbf{ 35.88 } }&\shortstack{ \\ 759.63 \\ \textbf{ 138.53 } }\\
\hline
fashion2000&\shortstack{ \\ 3.36 \\ \textbf{ 3.21 } }&\shortstack{ \\ 2.57 \\ \textbf{ 2.74 } }&\shortstack{ \\ 0.05 \\ \textbf{ 0.05 } }&\shortstack{ \\ 0.13 \\ \textbf{ 0.11 } }&\shortstack{ \\ 120.67 \\ \textbf{ 30.81 } }&\shortstack{ \\ 140.68 \\ \textbf{ 33.64 } }\\
\hline
fashion10000&\shortstack{ \\ 3.49 \\ \textbf{ 3.49 } }&\shortstack{ \\ 2.34 \\ \textbf{ 2.56 } }&\shortstack{ \\ 0.05 \\ \textbf{ 0.05 } }&\shortstack{ \\ 0.15 \\ \textbf{ 0.13 } }&\shortstack{ \\ 585.49 \\ \textbf{ 147.10 } }&\shortstack{ \\ 689.86 \\ \textbf{ 162.06 } }\\
\hline
\end{tabular}

}
\caption{Clustering scores: DB -- Davies Bouldin (\emph{lower}=\emph{better}) , SLH -- Silhouette Coefficient ($-1$: \emph{bad}, $1$: \emph{perfect}), CH - Calinski Harabasz (\emph{higher}=\emph{better}). Scores were computed as averages during  5-fold CV for training folds (normal font) and test folds (bold font). Worse values obtained after data transformation are marked with bullets.}
\label{tab:clustering_scores}
\end{table}

\subsection{FCM as preprocessing step}
During the last experiment we evaluated performance of multiple pipelines, each comprising three steps correspond to activities \emph{4.0}, \emph{7.0} and \emph{8.0} in Fig.~\ref{fig:experiments_design}:
\begin{enumerate}
\item Data preprocessing with a previously trained FCM classifier. During this step data in the original feature space $X$ was transformed to $X^{(d-1)}$ 
\item Training of selected classifier on data in $X^{(d-1)}$ 
\item Classifier evaluation
\end{enumerate}

Table~\ref{tab:preprocess_accuracy} and Table~\ref{tab:preprocess_f1} show mean accuracy and F1 scores obtained during 5-fold CV for datasets used in experiments. To facilitate comparison, we include pairs of scores $(cls,fcm+cls)$, where $cls$ stands for a classifier trained on original data and $fcm+cls$ corresponds to a pipeline, in which data are first preprocessed with trained FCM classifier and then submitted to train $cls$.
Please note that values in table are rounded, so sometimes one of identical values is indicated as higher.  

\begin{table}[ht!]
\centering
{\tiny

\begin{tabular}{||l||p{.63cm}|p{.63cm}||p{.63cm}|p{.63cm}||p{.63cm}|p{.63cm}||p{.63cm}|p{.63cm}||p{.63cm}|p{.63cm}||p{.63cm}|p{.63cm}||}
\hline
&mnb&fcm mnb&gnb&fcm gnb&knn3&fcm knn3&knn5&fcm knn5&svcrbf&fcm svcrbf&svclin&fcm svclin\\
\hline
iris&$0.77$&$\mathbf{ 0.96}$&$0.95$&$\mathbf{ 0.96}$&$\mathbf{ 0.95}$&$0.94$&$\mathbf{ 0.96}$&$\mathbf{ 0.96}$&$0.97$&$\mathbf{ 0.97}$&$0.97$&$\mathbf{ 0.97}$\\
wine&$0.93$&$\mathbf{ 0.97}$&$0.96$&$\mathbf{ 0.97}$&$0.95$&$\mathbf{ 0.97}$&$0.96$&$\mathbf{ 0.97}$&$0.97$&$\mathbf{ 0.98}$&$0.97$&$\mathbf{ 0.98}$\\
brest cancer&$0.84$&$\mathbf{ 0.96}$&$0.93$&$\mathbf{ 0.96}$&$0.96$&$\mathbf{ 0.97}$&$0.97$&$\mathbf{ 0.97}$&$0.97$&$\mathbf{ 0.97}$&$\mathbf{ 0.98}$&$0.97$\\
glass&$0.43$&$\mathbf{ 0.59}$&$0.32$&$\mathbf{ 0.44}$&$0.64$&$\mathbf{ 0.65}$&$\mathbf{ 0.64}$&$0.63$&$0.57$&$\mathbf{ 0.63}$&$0.54$&$\mathbf{ 0.64}$\\
seeds&$0.89$&$\mathbf{ 0.93}$&$0.89$&$\mathbf{ 0.91}$&$0.91$&$\mathbf{ 0.94}$&$0.92$&$\mathbf{ 0.93}$&$\mathbf{ 0.93}$&$0.93$&$\mathbf{ 0.93}$&$\mathbf{ 0.93}$\\
ionosphere&$0.66$&$\mathbf{ 0.84}$&$0.87$&$\mathbf{ 0.89}$&$0.83$&$\mathbf{ 0.89}$&$0.83$&$\mathbf{ 0.87}$&$0.90$&$\mathbf{ 0.91}$&$0.87$&$\mathbf{ 0.91}$\\
sonar&$\mathbf{ 0.66}$&$0.63$&$0.64$&$\mathbf{ 0.64}$&$0.59$&$\mathbf{ 0.63}$&$0.54$&$\mathbf{ 0.64}$&$0.61$&$\mathbf{ 0.67}$&$0.63$&$\mathbf{ 0.65}$\\
blood transfusion&$0.76$&$\mathbf{ 0.76}$&$\mathbf{ 0.75}$&$0.68$&$0.63$&$\mathbf{ 0.68}$&$0.65$&$\mathbf{ 0.67}$&$\mathbf{ 0.76}$&$0.76$&$\mathbf{ 0.76}$&$0.74$\\
vehicle&$0.53$&$\mathbf{ 0.74}$&$0.45$&$\mathbf{ 0.72}$&$0.70$&$\mathbf{ 0.75}$&$0.69$&$\mathbf{ 0.75}$&$0.71$&$\mathbf{ 0.77}$&$0.71$&$\mathbf{ 0.76}$\\
ecoli&$0.44$&$\mathbf{ 0.76}$&$0.72$&$\mathbf{ 0.76}$&$\mathbf{ 0.79}$&$0.78$&$0.78$&$\mathbf{ 0.78}$&$0.80$&$\mathbf{ 0.80}$&$0.80$&$\mathbf{ 0.80}$\\
yeast&$0.33$&$\mathbf{ 0.50}$&$0.14$&$\mathbf{ 0.20}$&$\mathbf{ 0.50}$&$0.46$&$\mathbf{ 0.53}$&$0.47$&$\mathbf{ 0.53}$&$0.50$&$\mathbf{ 0.53}$&$0.51$\\
tic tac toe&$0.67$&$\mathbf{ 0.75}$&$0.67$&$\mathbf{ 0.70}$&$\mathbf{ 0.76}$&$0.76$&$\mathbf{ 0.81}$&$0.75$&$0.73$&$\mathbf{ 0.75}$&$0.65$&$\mathbf{ 0.75}$\\
heart&$\mathbf{ 0.82}$&$0.81$&$\mathbf{ 0.84}$&$0.80$&$\mathbf{ 0.80}$&$0.77$&$\mathbf{ 0.81}$&$0.80$&$\mathbf{ 0.83}$&$0.81$&$\mathbf{ 0.85}$&$0.80$\\
haberman&$\mathbf{ 0.74}$&$\mathbf{ 0.74}$&$0.74$&$\mathbf{ 0.75}$&$0.69$&$\mathbf{ 0.71}$&$0.68$&$\mathbf{ 0.71}$&$0.73$&$\mathbf{ 0.73}$&$\mathbf{ 0.74}$&$\mathbf{ 0.74}$\\
german credit&$0.70$&$\mathbf{ 0.70}$&$\mathbf{ 0.70}$&$0.67$&$\mathbf{ 0.69}$&$0.68$&$\mathbf{ 0.71}$&$0.68$&$0.71$&$\mathbf{ 0.71}$&$0.69$&$\mathbf{ 0.70}$\\
diabets&$0.65$&$\mathbf{ 0.77}$&$0.75$&$\mathbf{ 0.76}$&$\mathbf{ 0.74}$&$0.74$&$0.74$&$\mathbf{ 0.74}$&$0.77$&$\mathbf{ 0.78}$&$0.77$&$\mathbf{ 0.78}$\\
olivetti faces 8&$0.76$&$\mathbf{ 0.92}$&$0.82$&$\mathbf{ 0.86}$&$0.85$&$\mathbf{ 0.91}$&$0.79$&$\mathbf{ 0.89}$&$0.70$&$\mathbf{ 0.92}$&$0.95$&$\mathbf{ 0.96}$\\
olivetti faces 16&$0.83$&$\mathbf{ 0.93}$&$0.85$&$\mathbf{ 0.92}$&$0.86$&$\mathbf{ 0.94}$&$0.81$&$\mathbf{ 0.92}$&$0.81$&$\mathbf{ 0.94}$&$0.97$&$\mathbf{ 0.98}$\\
olivetti faces 28&$0.86$&$\mathbf{ 0.94}$&$0.88$&$\mathbf{ 0.89}$&$0.90$&$\mathbf{ 0.93}$&$0.83$&$\mathbf{ 0.92}$&$0.85$&$\mathbf{ 0.90}$&$\mathbf{ 0.98}$&$0.97$\\
digits&$0.87$&$\mathbf{ 0.94}$&$0.80$&$\mathbf{ 0.92}$&$\mathbf{ 0.97}$&$0.95$&$\mathbf{ 0.96}$&$0.95$&$\mathbf{ 0.95}$&$0.95$&$\mathbf{ 0.95}$&$0.95$\\
fashion10000&$0.67$&$\mathbf{ 0.84}$&$0.52$&$\mathbf{ 0.80}$&$0.82$&$\mathbf{ 0.87}$&$0.82$&$\mathbf{ 0.87}$&$0.84$&$\mathbf{ 0.87}$&$0.83$&$\mathbf{ 0.86}$\\
\hline
\end{tabular}

\begin{tabular}{||l||p{.63cm}|p{.63cm}||p{.63cm}|p{.63cm}||p{.63cm}|p{.63cm}||p{.63cm}|p{.63cm}||p{.63cm}|p{.63cm}||p{.63cm}|p{.63cm}||}
\hline
&logreg&fcm logreg&dtree&fcm dtree&rforest&fcm rforest&gmc 1f&fcm gmc 1f&gmc 3f&fcm gmc 3f&gmc 3d&fcm gmc 3d\\
\hline
iris&$0.84$&$\mathbf{ 0.97}$&$\mathbf{ 0.96}$&$0.95$&$\mathbf{ 0.97}$&$0.95$&$\mathbf{ 0.98}$&$0.96$&$\mathbf{ 0.97}$&$0.96$&$\mathbf{ 0.96}$&$0.95$\\
wine&$0.97$&$\mathbf{ 0.98}$&$0.85$&$\mathbf{ 0.97}$&$\mathbf{ 0.98}$&$0.98$&$0.96$&$\mathbf{ 0.98}$&$0.86$&$\mathbf{ 0.98}$&$0.96$&$\mathbf{ 0.98}$\\
brest cancer&$0.96$&$\mathbf{ 0.97}$&$0.92$&$\mathbf{ 0.97}$&$0.96$&$\mathbf{ 0.96}$&$0.96$&$\mathbf{ 0.96}$&$0.95$&$\mathbf{ 0.97}$&$0.93$&$\mathbf{ 0.96}$\\
glass&$0.48$&$\mathbf{ 0.60}$&$\mathbf{ 0.62}$&$0.57$&$\mathbf{ 0.66}$&$0.64$&$0.50$&$\mathbf{ 0.51}$&$0.52$&$\mathbf{ 0.55}$&$0.46$&$\mathbf{ 0.59}$\\
seeds&$0.91$&$\mathbf{ 0.92}$&$\mathbf{ 0.90}$&$0.90$&$0.90$&$\mathbf{ 0.93}$&$\mathbf{ 0.93}$&$0.91$&$0.90$&$\mathbf{ 0.91}$&$0.89$&$\mathbf{ 0.90}$\\
ionosphere&$0.85$&$\mathbf{ 0.89}$&$0.86$&$\mathbf{ 0.87}$&$0.91$&$\mathbf{ 0.91}$&$0.89$&$\mathbf{ 0.91}$&$0.81$&$\mathbf{ 0.82}$&$0.88$&$\mathbf{ 0.89}$\\
sonar&$0.65$&$\mathbf{ 0.67}$&$0.61$&$\mathbf{ 0.63}$&$\mathbf{ 0.67}$&$0.64$&$0.59$&$\mathbf{ 0.62}$&$0.55$&$\mathbf{ 0.56}$&$0.56$&$\mathbf{ 0.62}$\\
blood transfusion&$\mathbf{ 0.77}$&$0.76$&$0.62$&$\mathbf{ 0.64}$&$0.66$&$\mathbf{ 0.68}$&$\mathbf{ 0.65}$&$0.63$&$\mathbf{ 0.63}$&$0.58$&$\mathbf{ 0.61}$&$0.59$\\
vehicle&$0.67$&$\mathbf{ 0.76}$&$0.72$&$\mathbf{ 0.74}$&$0.74$&$\mathbf{ 0.76}$&$\mathbf{ 0.83}$&$0.75$&$\mathbf{ 0.80}$&$0.74$&$0.67$&$\mathbf{ 0.74}$\\
ecoli&$0.75$&$\mathbf{ 0.80}$&$0.72$&$\mathbf{ 0.72}$&$0.79$&$\mathbf{ 0.80}$&$\mathbf{ 0.74}$&$0.73$&$\mathbf{ 0.74}$&$0.70$&$\mathbf{ 0.76}$&$0.72$\\
yeast&$\mathbf{ 0.53}$&$0.52$&$0.43$&$\mathbf{ 0.45}$&$\mathbf{ 0.52}$&$0.50$&$0.19$&$\mathbf{ 0.29}$&$0.41$&$\mathbf{ 0.43}$&$\mathbf{ 0.38}$&$0.31$\\
tic tac toe&$0.67$&$\mathbf{ 0.75}$&$\mathbf{ 0.76}$&$0.75$&$\mathbf{ 0.75}$&$0.75$&$0.56$&$\mathbf{ 0.73}$&$0.70$&$\mathbf{ 0.75}$&$0.58$&$\mathbf{ 0.75}$\\
heart&$\mathbf{ 0.83}$&$0.81$&$0.72$&$\mathbf{ 0.75}$&$0.78$&$\mathbf{ 0.79}$&$\mathbf{ 0.82}$&$0.79$&$0.71$&$\mathbf{ 0.80}$&$\mathbf{ 0.75}$&$0.73$\\
haberman&$\mathbf{ 0.74}$&$0.73$&$0.72$&$\mathbf{ 0.73}$&$\mathbf{ 0.73}$&$0.70$&$\mathbf{ 0.76}$&$0.73$&$0.64$&$\mathbf{ 0.65}$&$0.64$&$\mathbf{ 0.66}$\\
german credit&$\mathbf{ 0.71}$&$0.70$&$\mathbf{ 0.65}$&$0.63$&$\mathbf{ 0.70}$&$0.70$&$0.64$&$\mathbf{ 0.65}$&$0.59$&$\mathbf{ 0.64}$&$0.57$&$\mathbf{ 0.59}$\\
diabets&$0.76$&$\mathbf{ 0.78}$&$\mathbf{ 0.70}$&$0.69$&$\mathbf{ 0.74}$&$0.74$&$0.74$&$\mathbf{ 0.75}$&$0.66$&$\mathbf{ 0.73}$&$0.66$&$\mathbf{ 0.70}$\\
olivetti faces 8&$0.92$&$\mathbf{ 0.95}$&$0.48$&$\mathbf{ 0.63}$&$0.77$&$\mathbf{ 0.86}$&$\mathbf{ 0.97}$&$0.96$&$\mathbf{ 0.97}$&$0.96$&$0.82$&$\mathbf{ 0.89}$\\
olivetti faces 16&$\mathbf{ 0.97}$&$0.96$&$\mathbf{ 0.60}$&$0.58$&$0.78$&$\mathbf{ 0.83}$&$0.97$&$\mathbf{ 0.97}$&$0.97$&$\mathbf{ 0.97}$&$0.85$&$\mathbf{ 0.92}$\\
olivetti faces 28&$\mathbf{ 0.98}$&$0.96$&$\mathbf{ 0.64}$&$0.55$&$\mathbf{ 0.81}$&$0.77$&$\mathbf{ 0.97}$&$0.95$&$\mathbf{ 0.97}$&$0.95$&$0.88$&$\mathbf{ 0.88}$\\
digits&$0.93$&$\mathbf{ 0.94}$&$0.78$&$\mathbf{ 0.90}$&$0.90$&$\mathbf{ 0.93}$&$\mathbf{ 0.92}$&$0.91$&$0.73$&$\mathbf{ 0.90}$&$0.86$&$\mathbf{ 0.92}$\\
fashion10000&$0.83$&$\mathbf{ 0.87}$&$0.75$&$\mathbf{ 0.80}$&$0.82$&$\mathbf{ 0.85}$&$0.67$&$\mathbf{ 0.80}$&$0.65$&$\mathbf{ 0.82}$&$0.62$&$\mathbf{ 0.81}$\\
\hline
\end{tabular}

}
\caption{Accuracy: comparison of classifier scores on original dataset and transformed with previously trained FCM classifier}
\label{tab:preprocess_accuracy}
\end{table}

\begin{table}[ht!]
\centering
{\tiny

\begin{tabular}{||l||p{.63cm}|p{.63cm}||p{.63cm}|p{.63cm}||p{.63cm}|p{.63cm}||p{.63cm}|p{.63cm}||p{.63cm}|p{.63cm}||p{.63cm}|p{.63cm}||}
\hline
&mnb&fcm mnb&gnb&fcm gnb&knn3&fcm knn3&knn5&fcm knn5&svcrbf&fcm svcrbf&svclin&fcm svclin\\
\hline
iris&$0.77$&$\mathbf{ 0.96}$&$0.95$&$\mathbf{ 0.96}$&$\mathbf{ 0.95}$&$0.94$&$\mathbf{ 0.96}$&$0.96$&$\mathbf{ 0.97}$&$0.97$&$\mathbf{ 0.97}$&$\mathbf{ 0.97}$\\
wine&$0.93$&$\mathbf{ 0.97}$&$0.96$&$\mathbf{ 0.97}$&$0.95$&$\mathbf{ 0.97}$&$0.96$&$\mathbf{ 0.97}$&$0.97$&$\mathbf{ 0.98}$&$0.97$&$\mathbf{ 0.98}$\\
brest cancer&$0.81$&$\mathbf{ 0.96}$&$0.92$&$\mathbf{ 0.96}$&$0.96$&$\mathbf{ 0.97}$&$0.96$&$\mathbf{ 0.97}$&$0.96$&$\mathbf{ 0.97}$&$\mathbf{ 0.97}$&$0.97$\\
glass&$0.24$&$\mathbf{ 0.50}$&$0.34$&$\mathbf{ 0.44}$&$\mathbf{ 0.55}$&$0.54$&$0.47$&$\mathbf{ 0.52}$&$0.35$&$\mathbf{ 0.49}$&$0.33$&$\mathbf{ 0.50}$\\
seeds&$0.89$&$\mathbf{ 0.93}$&$0.89$&$\mathbf{ 0.91}$&$0.91$&$\mathbf{ 0.94}$&$0.92$&$\mathbf{ 0.93}$&$\mathbf{ 0.93}$&$0.93$&$0.93$&$\mathbf{ 0.93}$\\
ionosphere&$0.44$&$\mathbf{ 0.80}$&$0.85$&$\mathbf{ 0.88}$&$0.79$&$\mathbf{ 0.87}$&$0.79$&$\mathbf{ 0.85}$&$0.88$&$\mathbf{ 0.89}$&$0.84$&$\mathbf{ 0.89}$\\
sonar&$\mathbf{ 0.65}$&$0.63$&$0.64$&$\mathbf{ 0.64}$&$0.58$&$\mathbf{ 0.62}$&$0.54$&$\mathbf{ 0.63}$&$0.61$&$\mathbf{ 0.67}$&$0.62$&$\mathbf{ 0.64}$\\
blood transfusion&$0.43$&$\mathbf{ 0.44}$&$\mathbf{ 0.55}$&$0.51$&$0.47$&$\mathbf{ 0.52}$&$0.47$&$\mathbf{ 0.51}$&$0.44$&$\mathbf{ 0.53}$&$0.43$&$\mathbf{ 0.47}$\\
vehicle&$0.52$&$\mathbf{ 0.73}$&$0.43$&$\mathbf{ 0.71}$&$0.70$&$\mathbf{ 0.75}$&$0.68$&$\mathbf{ 0.75}$&$0.69$&$\mathbf{ 0.77}$&$0.70$&$\mathbf{ 0.76}$\\
ecoli&$0.22$&$\mathbf{ 0.49}$&$0.46$&$\mathbf{ 0.56}$&$\mathbf{ 0.65}$&$0.63$&$0.62$&$\mathbf{ 0.62}$&$0.59$&$\mathbf{ 0.61}$&$0.59$&$\mathbf{ 0.62}$\\
yeast&$0.12$&$\mathbf{ 0.35}$&$0.24$&$\mathbf{ 0.24}$&$\mathbf{ 0.44}$&$0.37$&$\mathbf{ 0.47}$&$0.37$&$\mathbf{ 0.34}$&$0.32$&$0.34$&$\mathbf{ 0.42}$\\
tic tac toe&$0.51$&$\mathbf{ 0.72}$&$0.62$&$\mathbf{ 0.69}$&$0.73$&$\mathbf{ 0.73}$&$\mathbf{ 0.79}$&$0.73$&$0.68$&$\mathbf{ 0.73}$&$0.58$&$\mathbf{ 0.72}$\\
heart&$\mathbf{ 0.82}$&$0.80$&$\mathbf{ 0.84}$&$0.79$&$\mathbf{ 0.79}$&$0.77$&$\mathbf{ 0.81}$&$0.80$&$\mathbf{ 0.83}$&$0.80$&$\mathbf{ 0.85}$&$0.80$\\
haberman&$\mathbf{ 0.42}$&$\mathbf{ 0.42}$&$0.56$&$\mathbf{ 0.63}$&$0.54$&$\mathbf{ 0.57}$&$0.52$&$\mathbf{ 0.58}$&$0.42$&$\mathbf{ 0.48}$&$0.42$&$\mathbf{ 0.43}$\\
german credit&$0.41$&$\mathbf{ 0.45}$&$\mathbf{ 0.64}$&$0.62$&$0.59$&$\mathbf{ 0.60}$&$\mathbf{ 0.60}$&$0.59$&$0.46$&$\mathbf{ 0.60}$&$0.43$&$\mathbf{ 0.61}$\\
diabets&$0.39$&$\mathbf{ 0.72}$&$0.72$&$\mathbf{ 0.74}$&$0.71$&$\mathbf{ 0.71}$&$0.70$&$\mathbf{ 0.71}$&$0.73$&$\mathbf{ 0.74}$&$0.73$&$\mathbf{ 0.75}$\\
olivetti faces 8&$0.75$&$\mathbf{ 0.91}$&$0.80$&$\mathbf{ 0.86}$&$0.84$&$\mathbf{ 0.91}$&$0.78$&$\mathbf{ 0.88}$&$0.66$&$\mathbf{ 0.91}$&$0.94$&$\mathbf{ 0.96}$\\
olivetti faces 16&$0.82$&$\mathbf{ 0.93}$&$0.84$&$\mathbf{ 0.91}$&$0.85$&$\mathbf{ 0.93}$&$0.80$&$\mathbf{ 0.91}$&$0.79$&$\mathbf{ 0.93}$&$0.97$&$\mathbf{ 0.98}$\\
olivetti faces 28&$0.86$&$\mathbf{ 0.93}$&$0.87$&$\mathbf{ 0.88}$&$0.89$&$\mathbf{ 0.92}$&$0.82$&$\mathbf{ 0.90}$&$0.84$&$\mathbf{ 0.90}$&$\mathbf{ 0.98}$&$0.96$\\
digits&$0.87$&$\mathbf{ 0.94}$&$0.80$&$\mathbf{ 0.92}$&$\mathbf{ 0.97}$&$0.95$&$\mathbf{ 0.96}$&$0.95$&$\mathbf{ 0.95}$&$0.95$&$\mathbf{ 0.95}$&$0.95$\\
fashion10000&$0.64$&$\mathbf{ 0.84}$&$0.47$&$\mathbf{ 0.80}$&$0.81$&$\mathbf{ 0.87}$&$0.81$&$\mathbf{ 0.87}$&$0.83$&$\mathbf{ 0.87}$&$0.83$&$\mathbf{ 0.86}$\\
\hline
\end{tabular}

\begin{tabular}{||l||p{.63cm}|p{.63cm}||p{.63cm}|p{.63cm}||p{.63cm}|p{.63cm}||p{.63cm}|p{.63cm}||p{.63cm}|p{.63cm}||p{.63cm}|p{.63cm}||}
\hline
&logreg&fcm logreg&dtree&fcm dtree&rforest&fcm rforest&gmc 1f&fcm gmc 1f&gmc 3f&fcm gmc 3f&gmc 3d&fcm gmc 3d\\
\hline
iris&$0.83$&$\mathbf{ 0.97}$&$\mathbf{ 0.96}$&$0.95$&$\mathbf{ 0.97}$&$0.95$&$\mathbf{ 0.98}$&$0.96$&$\mathbf{ 0.97}$&$0.96$&$\mathbf{ 0.96}$&$0.95$\\
wine&$0.97$&$\mathbf{ 0.98}$&$0.85$&$\mathbf{ 0.97}$&$\mathbf{ 0.98}$&$0.98$&$0.96$&$\mathbf{ 0.98}$&$0.86$&$\mathbf{ 0.98}$&$0.96$&$\mathbf{ 0.98}$\\
brest cancer&$0.96$&$\mathbf{ 0.97}$&$0.92$&$\mathbf{ 0.97}$&$0.96$&$\mathbf{ 0.96}$&$0.96$&$\mathbf{ 0.96}$&$0.95$&$\mathbf{ 0.96}$&$0.93$&$\mathbf{ 0.95}$\\
glass&$0.29$&$\mathbf{ 0.46}$&$\mathbf{ 0.54}$&$0.48$&$\mathbf{ 0.59}$&$0.52$&$\mathbf{ 0.43}$&$0.33$&$\mathbf{ 0.48}$&$0.43$&$0.41$&$\mathbf{ 0.49}$\\
seeds&$0.91$&$\mathbf{ 0.92}$&$\mathbf{ 0.90}$&$0.90$&$0.90$&$\mathbf{ 0.93}$&$\mathbf{ 0.93}$&$0.91$&$0.90$&$\mathbf{ 0.92}$&$0.89$&$\mathbf{ 0.91}$\\
ionosphere&$0.81$&$\mathbf{ 0.87}$&$0.85$&$\mathbf{ 0.86}$&$0.89$&$\mathbf{ 0.90}$&$0.89$&$\mathbf{ 0.90}$&$0.80$&$\mathbf{ 0.82}$&$0.88$&$\mathbf{ 0.88}$\\
sonar&$0.65$&$\mathbf{ 0.67}$&$0.60$&$\mathbf{ 0.63}$&$\mathbf{ 0.67}$&$0.63$&$0.56$&$\mathbf{ 0.61}$&$0.48$&$\mathbf{ 0.49}$&$0.55$&$\mathbf{ 0.61}$\\
blood transfusion&$0.49$&$\mathbf{ 0.54}$&$0.48$&$\mathbf{ 0.51}$&$0.51$&$\mathbf{ 0.51}$&$\mathbf{ 0.59}$&$0.52$&$\mathbf{ 0.57}$&$0.50$&$\mathbf{ 0.53}$&$0.52$\\
vehicle&$0.65$&$\mathbf{ 0.76}$&$0.72$&$\mathbf{ 0.74}$&$0.73$&$\mathbf{ 0.76}$&$\mathbf{ 0.83}$&$0.75$&$\mathbf{ 0.80}$&$0.74$&$0.67$&$\mathbf{ 0.74}$\\
ecoli&$0.49$&$\mathbf{ 0.62}$&$0.46$&$\mathbf{ 0.67}$&$\mathbf{ 0.65}$&$0.64$&$0.49$&$\mathbf{ 0.60}$&$\mathbf{ 0.60}$&$0.48$&$\mathbf{ 0.59}$&$0.55$\\
yeast&$0.29$&$\mathbf{ 0.39}$&$0.28$&$\mathbf{ 0.36}$&$0.37$&$\mathbf{ 0.42}$&$0.21$&$\mathbf{ 0.28}$&$0.36$&$\mathbf{ 0.37}$&$\mathbf{ 0.36}$&$0.29$\\
tic tac toe&$0.61$&$\mathbf{ 0.73}$&$\mathbf{ 0.73}$&$0.72$&$0.72$&$\mathbf{ 0.72}$&$0.52$&$\mathbf{ 0.71}$&$0.68$&$\mathbf{ 0.74}$&$0.57$&$\mathbf{ 0.73}$\\
heart&$\mathbf{ 0.83}$&$0.80$&$0.72$&$\mathbf{ 0.75}$&$0.78$&$\mathbf{ 0.79}$&$\mathbf{ 0.82}$&$0.78$&$0.71$&$\mathbf{ 0.80}$&$\mathbf{ 0.75}$&$0.72$\\
haberman&$0.45$&$\mathbf{ 0.52}$&$0.56$&$\mathbf{ 0.60}$&$\mathbf{ 0.58}$&$0.56$&$\mathbf{ 0.63}$&$0.62$&$\mathbf{ 0.58}$&$0.58$&$0.60$&$\mathbf{ 0.60}$\\
german credit&$0.56$&$\mathbf{ 0.60}$&$\mathbf{ 0.60}$&$0.56$&$\mathbf{ 0.57}$&$0.57$&$\mathbf{ 0.61}$&$0.60$&$0.53$&$\mathbf{ 0.57}$&$0.52$&$\mathbf{ 0.55}$\\
diabets&$0.71$&$\mathbf{ 0.75}$&$\mathbf{ 0.67}$&$0.66$&$\mathbf{ 0.70}$&$0.70$&$0.71$&$\mathbf{ 0.73}$&$0.63$&$\mathbf{ 0.71}$&$0.66$&$\mathbf{ 0.69}$\\
olivetti faces 8&$0.91$&$\mathbf{ 0.94}$&$0.45$&$\mathbf{ 0.60}$&$0.75$&$\mathbf{ 0.84}$&$\mathbf{ 0.97}$&$0.96$&$\mathbf{ 0.97}$&$0.96$&$0.80$&$\mathbf{ 0.88}$\\
olivetti faces 16&$\mathbf{ 0.97}$&$0.96$&$0.56$&$\mathbf{ 0.57}$&$0.76$&$\mathbf{ 0.81}$&$0.97$&$\mathbf{ 0.97}$&$0.97$&$\mathbf{ 0.97}$&$0.84$&$\mathbf{ 0.90}$\\
olivetti faces 28&$\mathbf{ 0.97}$&$0.96$&$\mathbf{ 0.62}$&$0.52$&$\mathbf{ 0.80}$&$0.76$&$\mathbf{ 0.97}$&$0.95$&$\mathbf{ 0.97}$&$0.95$&$0.87$&$\mathbf{ 0.87}$\\
digits&$0.93$&$\mathbf{ 0.94}$&$0.78$&$\mathbf{ 0.90}$&$0.90$&$\mathbf{ 0.93}$&$\mathbf{ 0.92}$&$0.91$&$0.73$&$\mathbf{ 0.91}$&$0.86$&$\mathbf{ 0.92}$\\
fashion10000&$0.83$&$\mathbf{ 0.86}$&$0.75$&$\mathbf{ 0.79}$&$0.82$&$\mathbf{ 0.85}$&$0.64$&$\mathbf{ 0.80}$&$0.63$&$\mathbf{ 0.82}$&$0.61$&$\mathbf{ 0.81}$\\
\hline
\end{tabular}

}
\caption{F1 macro: comparison of classifier scores on original dataset and transformed with previously trained FCM classifier. }
\label{tab:preprocess_f1}
\end{table}


We performed statistical evaluation of results for $(cls,fcm+cls)$, during which two tests: matched samples t-test and Wilcoxon signed ranks were applied. The t-test assumes null hypothesis that difference between mean performance values is not significant. The second, Wilcoxon test is based on difference between classifier performance converted into ranks. The null hypothesis is that they are symmetrically distributed around 0, i.e. not shifted. Opposite situation would indicate superiority of one of the classifiers \cite{japkowicz2011evaluating}.   

The test were performed with respect to accuracy and F1 macro values. Two pairs of samples were used: mean score values obtained for 21 datasets during 5-fold CV and results for folds data (105 samples). Please recall that exactly the same folds were used in all experiments (c.f. Fig.~\ref{fig:experiments_design}). Table~\ref{tab:pairs_stats-cls_vs_fcm_cls} shows the summary results. If a test was not capable of rejecting the null hypothesis, equal sign is put for both configurations. In the opposite case plus (minus) signs are used to mark the better (worse) classifier.  We set significance level $\alpha = 0.05$.

\begin{table}[ht!]
\centering
{\tiny

\begin{tabular}{|l|l||c|c|c|c|c|c|c|c||c|c|c|c|c|c|c|c|}
\hline
Alg&FCM+Alg&\multicolumn{8}{c|}{Accuracy}&\multicolumn{8}{c|}{F1 macro}\\
\cline{3-10}\cline{11-18}
&&\multicolumn{4}{c|}{datasets}&\multicolumn{4}{c|}{folds}&\multicolumn{4}{c|}{datasets}&\multicolumn{4}{c|}{folds}\\
\cline{3-10}\cline{11-18}
&&\multicolumn{2}{c|}{t-test}&\multicolumn{2}{c|}{Wilcoxon}&\multicolumn{2}{c|}{t-test}&\multicolumn{2}{c|}{Wilcoxon}&\multicolumn{2}{c|}{t-test}&\multicolumn{2}{c|}{Wilcoxon}&\multicolumn{2}{c|}{t-test}&\multicolumn{2}{c|}{Wilcoxon}\\

\hline
mnb&fcm mnb&-&+&-&+&-&+&-&+&-&+&-&+&-&+&-&+\\
gnb&fcm gnb&-&+&-&+&-&+&-&+&-&+&-&+&-&+&-&+\\
knn3&fcm knn3&-&+&-&+&-&+&-&+&-&+&-&+&-&+&-&+\\
knn5&fcm knn5&-&+&=&=&-&+&-&+&-&+&-&+&-&+&-&+\\
svcrbf&fcm svcrbf&-&+&-&+&-&+&-&+&-&+&-&+&-&+&-&+\\
svclin&fcm svclin&=&=&=&=&-&+&-&+&-&+&-&+&-&+&-&+\\
logreg&fcm logreg&-&+&-&+&-&+&-&+&-&+&-&+&-&+&-&+\\
dtree&fcm dtree&=&=&=&=&-&+&-&+&-&+&-&+&-&+&-&+\\
rforest&fcm rforest&=&=&=&=&=&=&=&=&=&=&=&=&=&=&=&=\\
gmc 1f&fcm gmc 1f&=&=&=&=&=&=&=&=&=&=&=&=&=&=&=&=\\
gmc 3f&fcm gmc 3f&-&+&-&+&-&+&-&+&=&=&=&=&-&+&-&+\\
gmc 3d&fcm gmc 3d&-&+&-&+&-&+&-&+&-&+&-&+&-&+&-&+\\
\hline
\end{tabular}

}
\caption{Statistical comparison of classifier performance tested on original data and  data preprocessed with FCM. The table shows results of paired t-test and Wilcoxon for mean score values for the whole datasets, as well as particular folds during 5-fold CV.}
\label{tab:pairs_stats-cls_vs_fcm_cls}
\end{table}

Results returned by t-test are consistent with Wilcoxon test (with one exception in the case of \emph{knn5} (k-Nearest Neighbors, k=5) classifier. 
As test sensitivity (i.e. ability to reject null hypothesis) usually grows with sample size, comparisons made based on folds are probably more accurate. Focusing on results obtained for folds, it can be noticed that results for accuracy and F1 macro are exactly the same. Moreover, in all cases apart from \emph{rforest} (Random Forest) and \emph{gmc 1f} (Gaussian Mixture with 1 component and full covariance matrix) results for pipelines consisting of FCM transformer and a classifier were significantly better than for the classifier applied alone.

The fact that preprocessing with trained FCM classifier yielded better results for Na\"ive Bayes and Gaussian Mixture (which relay on data compactness),
as well as SVM and Logistic Regression (which in turn rather relay on separability) confirms the hypothesis that FCM classifier transforms the feature space so that both dataset properties improve.


Another statistical evaluation aimed at answering the question, \emph{whether combination of FCM transformer and a classifier outperforms the FCM based classifier}.
If the answer was negative, there would be no justification for building a pipeline by first training an FCM and then another classifier on transformed data.  

The conditions were exactly the same as in the previous tests. Results  gathered in Table~\ref{tab:pairs_stats-fcm_vs_fcm_cls} indicate that application of pipeline consisting of FCM+classifier can be profitable only for three configurations: \emph{svclin} (SVM with linear kernel), \emph{svcrbf} (SVM with RBF kernel) and \emph{logreg} Logistic Regression. 

After a moment of reflection these results turn out to be not surprising. SVM is strongly based on the concept of data separability. Preprocessing done within $d-1$ FCM iterations seems to make the data more separable, however, its last $d$-th classification step, which  corresponds to the very basic Logistic Regression, is probably less efficient than hyperplane selection combined with soft margin optimization done by SVM. The same applies to the Logistic Regression: its library implementation is more mature, e.g. by default applies regularization.

\begin{table}[ht!]
\centering
{\tiny

\begin{tabular}{|l|l||c|c|c|c|c|c|c|c||c|c|c|c|c|c|c|c|}
\hline
Alg&FCM+Alg&\multicolumn{8}{c|}{Accuracy}&\multicolumn{8}{c|}{F1 macro}\\
\cline{3-10}\cline{11-18}
&&\multicolumn{4}{c|}{datasets}&\multicolumn{4}{c|}{folds}&\multicolumn{4}{c|}{datasets}&\multicolumn{4}{c|}{folds}\\
\cline{3-10}\cline{11-18}
&&\multicolumn{2}{c|}{t-test}&\multicolumn{2}{c|}{Wilcoxon}&\multicolumn{2}{c|}{t-test}&\multicolumn{2}{c|}{Wilcoxon}&\multicolumn{2}{c|}{t-test}&\multicolumn{2}{c|}{Wilcoxon}&\multicolumn{2}{c|}{t-test}&\multicolumn{2}{c|}{Wilcoxon}\\

\hline
fcm&fcm mnb&=&=&=&=&=&=&=&=&=&=&=&=&=&=&=&=\\
fcm&fcm gnb&=&=&+&-&+&-&+&-&=&=&=&=&=&=&=&=\\
fcm&fcm knn3&=&=&=&=&=&=&=&=&=&=&=&=&-&+&-&+\\
fcm&fcm knn5&=&=&=&=&=&=&=&=&=&=&=&=&-&+&=&=\\
fcm&fcm svcrbf&-&+&-&+&-&+&-&+&-&+&=&=&-&+&-&+\\
fcm&fcm svclin&-&+&-&+&-&+&-&+&=&=&=&=&-&+&-&+\\
fcm&fcm logreg&-&+&-&+&-&+&-&+&-&+&-&+&-&+&-&+\\
fcm&fcm dtree&+&-&+&-&+&-&+&-&=&=&+&-&+&-&+&-\\
fcm&fcm rforest&=&=&=&=&=&=&+&-&=&=&=&=&=&=&=&=\\
fcm&fcm gmc 1f&=&=&=&=&=&=&+&-&=&=&=&=&=&=&=&=\\
fcm&fcm gmc 3f&=&=&+&-&=&=&+&-&=&=&=&=&=&=&=&=\\
fcm&fcm gmc 3d&+&-&+&-&+&-&+&-&=&=&=&=&=&=&=&=\\
\hline
\end{tabular}

}
\caption{Statistical comparison of FCM vs FCM + classifier. The table shows results of paired t-test and Wilcoxon for mean score values for the whole datasets, as well as particular folds during 5-fold CV.}
\label{tab:pairs_stats-fcm_vs_fcm_cls}
\end{table}

\subsection{Discussion}
\label{subsec:discussion}

%
%
%

\paragraph{Comparison}

A characteristic feature of the proposed classifier, which distinguishes it from other FCM based solutions discussed in Section~\ref{subsec:classification}, 
is the fully connected map architecture allowing backward links and the lack of direct dependence on convergence to fixed point attractors. 

Analysis of simple cases, like these in Fig.~\ref{fig:ex_fcm_attractors} suggested that even a few steps of FCM execution is enough to push trajectories towards attractors. At the same time, the final states of trajectories become more and more condensed, what facilitates separating their groups. 

The convergence requirement was rejected after we realized, that it would have been very hard to implement, especially for  higher dimensions. First, the convergence would have to be properly detected and distinguished from cyclic and chaotic behavior. Second, as multiple iterations might be required to converge, the whole process would be computationally expensive. 

Moreover, our intention was to use up to date gradient optimization techniques, which due to vanishing  gradient problem would probably occur ineffective for multiple iterations. It may be observed, that gradient methods were applied principally to time series prediction occurring within a few steps, see for example \cite{papageorgiou2017fuzzy}. The methods based on a convergence to a fixed point attractors used other optimization techniques, e.g. a genetic algorithm for map learning \cite{froelich2017towards}.

%

\paragraph{The weights matrix}

Typically, the elements of FCM weights matrix designed by experts have values from the interval  $[-1,1]$.
The described FCM learning procedure does not put constraints on weights, in consequence they usually exceed these bounds.  However, each row of $W$ and corresponding $b$ element can be normalized to fit within $[-1,1]$. Let  $s_i=\max(\{W_{ij}\}_{j=1,r}\cup\{b_i\})$ The state equation (\ref{eq:state-eq-classifier}) for $i$-th concept can be rewritten as:

\begin{multline}
A^{(t)}_i=f(W_i\cdot A^{(t-1)}+b_i)=f(s_i\cdot(\frac{W_i}{s_i} A^{(t-1)}+\frac{b_i}{s_i})\\
=\frac{1}{1+\exp(\lambda\cdot s_i(\frac{W_i}{s_i} A^{(t-1)}+\frac{b_i}{s_i}))}
\label{eq:state_equation_normalized}
\end{multline}

The formula (\ref{eq:state_equation_normalized}) can be interpreted as employment of individual activation functions $f_i$ for each concept, each with different slope coefficient $\lambda_i=\lambda\cdot s_i$. 

At this point a question arises, why to use $\lambda$ as the model parameter? Technically, regardless of  $\lambda$ values, the learning algorithm may find weights appropriately scaled, so $\lambda$ might be set to 1 or simply omitted. This is true for majority of configurations given in Table~\ref{tab:fcm_params}. However, in some situations selecting $\lambda \neq 1$ improves the learning process. During gradient computation, either with GradientTape or Algorithm~\ref{alg:back_prop}, the activation function $f(x)$ is differentiated $d$ times. 
As its derivative is equal $f(x)'=\lambda f(x)(1-f(x)$,  the final gradient comprises terms scaled by powers of $\lambda$: $\lambda^1,\lambda^2,\dots,\lambda^d$. Setting $\lambda >1$ prevents diminishing gradients, whereas $\lambda < 1$ allows to avoid gradient explosion. 


\paragraph{Parameters}
Classification algorithms come with a number of parameters that either are part of the model or are used to control the learning process. Mature implementations often define their default values or have built in procedures that set them based on data used in learning.  

As the implementation of FCM classifier is fresh, we cannot offer good default parameters. 
As can be observed in Table~\ref{tab:fcm_params} we often used depth $d=3$ as an initial guess. Selection of $bs=-1$ (no batches) in the majority of cases was done due to learning performance. Number of epochs was usually set experimentally, by monitoring loss during learning (this at least can be easily automated).

Typically, small values of $\lambda$ from the range $[1,3]$ were used. As derivative of a sigmoid function is $f'(x)=\lambda f(x)(1-f(x))$, c.f. equation (\ref{eq:sigmoid_derivative}), the $\lambda$ parameter and the learning rate $lr$ interfere and their product affected strongly the learning process. This in particular can be observed in the case of \emph{rmsprop} optimization algorithm. For \emph{adam}, the default learning rate, which is 0.001, yielded good results.


\section{Conclusions}
\label{sec:conclusion}

In this paper we proposed a classification method based on Fuzzy Cognitve Maps. 
It employed a fully connected map structure allowing links between all types of concepts. Designing the classifier, we assumed that reasoning would be performed during a fixed number of FCM iterations. We applied a gradient algorithm for model learning, which in the prototype software was realized by means of symbolic differentiation. However, more effcient differentiation method was also described.

The performance of FCM based classifier turned out to be competitive with classical machine learning algorithms, what was proven by statistical tests. Aware of the No Free Lunch theorem, we were not expecting to develop an algorithm that for all datasets would outperform other classification methods. However, it can be stated that the FCM classifier belongs to the group of algorithms that attained the highest performance scores.

We used a prototype implementation based on relatively slow symbolic differentiation offered by GradientTape mechanism of TensorFlow. This limits the size of datasets that could be processed in acceptable time. The proposed method was tested on a group of relatively small datasets, the largest  \emph{fashion10000} comprised 784 attributes and 10000 instances. 


The hypothesis that FCM classifier is capable of transforming feature space so that observations belonging to a given class become more condensed and easier to separate was confirmed by two tests consisting in calculating internal clustering scores and building pipelines composed of FCM transformer and a classification algorithm. 
We consider this resultant important as it opens opportunity to investigate various hybrid  architectures combining FCM and other classification algorithms.

For the next future we plan to replace symbolic gradient calculation by the hard coded backpropagation algorithm, preferably, redesigning it as a component compatible with the Keras library. We expect that this  will significantly reduce the time of training. 

Another path, which would probably allow to improve time efficiency of the learning phase, is weights matrix initialization, e.g. based on correlation between attributes and class labels.  

As the size of model grows quadratically with number of data attributes, FCM based classifier may be hard to learn for high-dimensional data. To alleviate this problem we plan to combine it with known  dimensionality reduction methods.





\bibliographystyle{model1-num-names}
\bibliography{pszwed,FCM,refs}







\end{document}